\documentclass{article}

\usepackage{microtype}
\usepackage{graphicx}
\usepackage{subfigure}
\usepackage{booktabs} 

\usepackage{hyperref}

\usepackage[accepted]{icml2024}
\usepackage{amsmath}
\usepackage{amssymb}
\usepackage{mathtools}
\usepackage{amsthm}

\usepackage[capitalize,noabbrev]{cleveref}

\theoremstyle{plain}

\usepackage{cuted}

\usepackage{tcolorbox}

\newcommand{\smallsquare}{\scalebox{0.35}{$\blacksquare$}} 

\definecolor{mycustomcolor}{rgb}{0.196, 0.349, 0.490}
\definecolor{myHexColor}{HTML}{2e598f}

\definecolor{softlightblue}{rgb}{0.8, 0.9, 0.97}
\definecolor{softlightgray}{rgb}{0.9, 0.9, 0.9}
\definecolor{lightred}{rgb}{1, 0.8, 0.8}

\definecolor{green_}{rgb}{0.0, 0.5, 0.0}  
\definecolor{red_}{rgb}{0.6, 0.0, 0.0}
\definecolor{blue_}{rgb}{0.0, 0.0, 0.8}
\definecolor{transpose_}{rgb}{0.05, 0.5, 0.86}

\definecolor{egyptianblue}{rgb}{0.06, 0.2, 0.65}
\definecolor{mediumpersianblue}{rgb}{0.0, 0.4, 0.65}

\definecolor{mycolor}{RGB}{123, 223, 242}

\definecolor{silver}{RGB}{218, 229, 236}

\definecolor{mycustomcolordark}{RGB}{98, 178, 194} 

\definecolor{bleudefrance}{rgb}{0.19, 0.55, 0.91}

\usepackage{tocbibind} 
\usepackage{appendix} 
\usepackage{hyperref}  

\usepackage{wrapfig}

\usepackage{graphicx}
\usepackage{array}
\usepackage{float}
\usepackage{caption}

\usepackage{tikz}
\usetikzlibrary{arrows.meta}  

 \usetikzlibrary{calc}

\usepackage{amsmath}
\usepackage{xcolor}

\definecolor{deepcarmine}{rgb}{0.61, 0.08, 0.19}
\definecolor{twilightlavender}{rgb}{0.54, 0.29}
\definecolor{upmaroon}{rgb}{0.48, 0.07, 0.07}

\newcounter{theoremBox}

\newtcolorbox[auto counter]{theorem}[2][]{
    colframe = deepcarmine!40, 
    coltitle=black, 
    title={Theorem \thetcbcounter:  #2},
    fonttitle=\bfseries,
    arc=1mm, 
    boxrule=0.5mm,
    top=2mm,
    bottom=2mm,
    left=2mm,
    right=2mm,
    before=\vspace{1ex}, 
    after=\vspace{1ex}, 
    code={\refstepcounter{theoremBox}\ifstrempty{#1}{}{\label{#1}}}
}
\newcounter{defbox}

\newtcolorbox[auto counter]{definition}[2][]{
    colback=teal!1, 
    colframe=mycustomcolor!60, 
    coltitle=black, 
    title={Definition \thetcbcounter: #2},
    fonttitle=\bfseries,
    arc=1mm, 
    boxrule=0.5mm,
    top=2mm,
    bottom=2mm,
    left=2mm,
    right=2mm,
    code={\refstepcounter{defbox}\ifstrempty{#1}{}{\label{#1}}}
}

\usepackage[capitalize,noabbrev]{cleveref}

\theoremstyle{plain}

\usepackage[textsize=tiny]{todonotes}

\icmltitlerunning{Adaptive Sampling for Continuous Group Equivariant Neural Networks}

\begin{document}

\twocolumn[
\icmltitle{Adaptive Sampling for Continuous Group Equivariant Neural Networks}



\icmlsetsymbol{equal}{*}
\begin{icmlauthorlist}
\icmlauthor{Berfin Inal}{quva,uva}
\icmlauthor{Gabriele Cesa}{comp,uva}
\end{icmlauthorlist}

\icmlaffiliation{quva}{Work done during internship in QUVA-Lab. All datasets were downloaded/generated in QUVA-Lab}
\icmlaffiliation{comp}{Qualcomm AI Research, an initiative of Qualcomm Technologies, Inc}
\icmlaffiliation{uva}{University of Amsterdam}

\icmlcorrespondingauthor{Berfin Inal}{s.berfininal@gmail.com}
\icmlcorrespondingauthor{Gabriele Cesa}{gcesa@qti.qualcomm.com}

\icmlkeywords{adaptive sampling, steerable networks, equivariant networks, equivariant nonlinearities, computational efficiency}

\vskip 0.3in
\editorsListText
\vskip 0.3in
]



\printAffiliationsAndNotice{}  






\begin{abstract}
Steerable networks, which process data with intrinsic symmetries, often use Fourier-based non-linearities that require sampling from the entire group, leading to a need for discretization in continuous groups. As the number of samples increases, both performance and equivariance improve, yet this also leads to higher computational costs. To address this, we introduce an adaptive sampling approach that dynamically adjusts the sampling process to the symmetries in the data, reducing the number of required group samples and lowering the computational demands. We explore various implementations and their effects on model performance, equivariance, and computational efficiency. Our findings demonstrate improved model performance, and a marginal increase in memory
efficiency. 
\end{abstract}

\section{Introduction}

Symmetry processing holds a significant importance in deep learning, since many real-world datasets inherently exhibit symmetrical properties. Research in integrating equivariance into deep architectures, such as steerable CNNs and Group Convolutional Neural Networks (GCNNs), has become a significant focus \cite{group_CNN, steerable_cnn, harmonic_networks, tensor_field_networks, 3D_steerable_cnns}. While traditional CNNs offer translation equivariance, the first GCNNs \cite{group_CNN} extended this to rotations and reflections, though  limited to small discrete

\begin{figure}[!h]
    \centering
\includegraphics[width = 1\columnwidth]{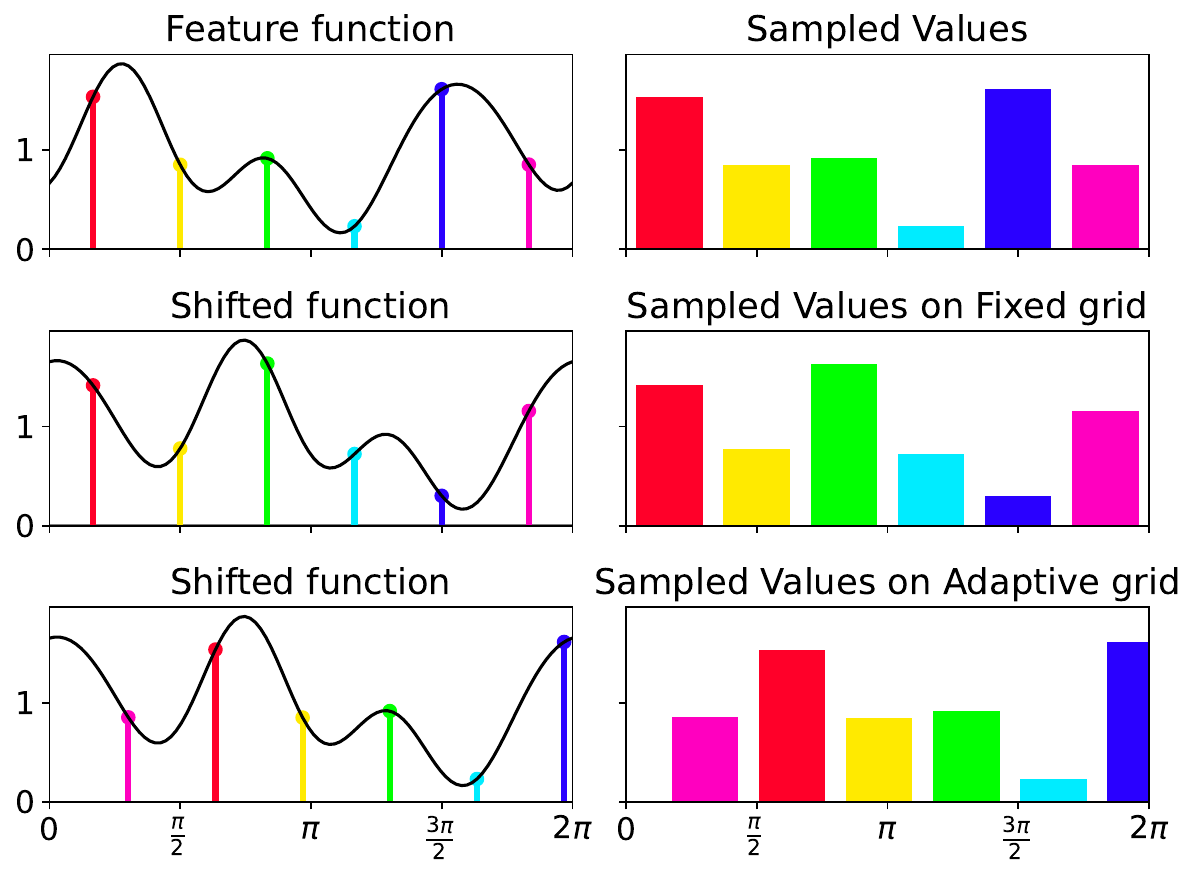}
    \caption{Behaviour of sampled values from a translated feature when using a fixed and an adaptive grid. \textit{Left column}: the continuous feature function as well as the location of the grid samples (in colors). \textit{Right column}: the measured values at the grid locations. When the function is shifted, the sampled values on the \textit{fixed grid} change; these samples can not accurately capture translations smaller than the grid resolution. Instead, the \textit{adaptive grid} is translated together with the input function, ensuring the measured values are constant (compare columns of the same color in the first and third row). Information about the phase is preserved by the adaptive grid itself (note the colored grid points translates too).} \label{fig:adaptive_grid_visualization}
\end{figure}

groups like $C_4$ and $D_4$ due to computational limits.

Steerable CNNs, introduced by \citet{steerable_cnn}, use steerable filters to achieve equivariance to larger groups with reduced computational demands. In the design of steerable CNNs, choosing the right activation function presents a key challenge. While pointwise nonlinearities have proven to be highly effective, they demand sampling from the entire group, requiring some form of discretization for continuous groups \cite{general_nonlinearities_on_so2_equivariant_cnns, equivariant_mesh_cnn, escnn}.
This introduces a trade-off between compute and performance:
using more samples in the discretization improves model stability and equivariance, yet it also raises the computational and memory cost of the layer.

In our work, we aim to improve the sampling process within these pointwise non-linearities by using an adaptive grid approach.
Traditionally, group sampling is done on a fixed grid, where samples are generated at model initialization and then cached for use throughout the network. 
However, our method dynamically \textit{adjusts the sampling grid to align with the input data}.
More precisely, we propose predicting the sampling grid from the input in an equivariant way: this ensures that transformed versions of the same input are processed in the non-linear layers using accordingly transformed versions of the same sampling grid.
As proven in Sec.~\ref{sec:adaptive sampling matrix}, this guarantees that the activation layer is always perfectly equivariant regardless of the number of samples employed.
Fig~\ref{fig:adaptive_grid_visualization} provides an intuitive visualization of this idea.

We present various implementations of the adaptive grid, focusing on strategies to share it efficiently and reduce the overall computational costs.
Our empirical analysis examines how varying the number of samples, compared to a fixed grid, affects model performance, equivariance error, and computational efficiency. 
Our findings indicate improved model performance and a marginal increase in memory efficiency. 
Lastly, we explore the limitations, computational considerations, and potential advancements of our approach.

\section{Steerable Convolutional Neural Networks}\label{sec:steerable cnns} 
\paragraph{Steerable Features}
In steerable CNNs, feature spaces are defined as spaces of \textit{steerable feature fields} $f : \mathbb{R}^n \rightarrow \mathbb{R}^{d_{\rho}}$, which assigns a $d_{\rho}$-dimensional vector $f(x) \in \mathbb{R}^{d_\rho}$ to each data point $x \in \mathbb{R}^n$. 
These feature fields are associated with a transformation law, which describes how they are transformed by the action of the group $G$. 
The transformation law of a $d_{\rho}$-dimensional vector is defined by a group representation $\rho : G \rightarrow \mathbb{R}^{d_\rho \times d_\rho}$:
\begin{align}
    [g.f](x) := \rho(g) f(g^{-1}.x)
\end{align}
where $\rho$ specifies how the $d_\rho$ channels of each feature vector $f(x)$ mix.
In practice, the entire feature space can be defined as the \textit{direct sum} $\bigoplus_i f_i$ of multiple individual feature fields $f_i$, which transforms according to the direct sum representation $\rho := \bigoplus_i \rho_i$, enabling each field to transform independently of others. 

An example of feature fields are scalar fields, which transform according to the trivial representation $\psi_0(g) = 1$. 
GCNNs are special cases of steerable CNNs using intermediate features that transform according to the \textit{regular representation} of $G$; see Sec.~\ref{sec: Regular and Quotient nonlinearities}.
Steerable CNNs generalize GCNNs by allowing feature fields to transform according to more complex geometric types.
This is achieved by the steerable kernel which satisfy the \textit{steerability constraint}. Given the feature fields with the input type $\rho_{\text{in}} : G \rightarrow \mathbb{R}^{d_{\text{in}} \times d_{\text{in}}}$ and the output type $\rho_\text{out} : G \rightarrow \mathbb{R}^{d_{\text{out}} \times d_{\text{out}}}$, the $G$-steerable convolutional kernel $K : \mathbb{R}^n \rightarrow \mathbb{R}^{d_{\text{out}} \times d_{\text{in}}}$ satisfies the steerability constraint $\forall g \in G, x \in \mathbb{R}^n$:
\begin{align}
K(x) = \rho_{\text{out}}(g)K(g^{-1}x)\rho
_{\text{in}}(g)^{-1}.
\end{align}

Any equivariant kernel can be described using a $G-$steerable kernel basis, as detailed in \citet{escnn}. 
For parametrization, these basis elements are combined with learnable weights. Since each pair of $\rho_{\text{in}}$ and $\rho_{\text{out}}$ requires a unique solution, the kernel constraint can be simplified by breaking the input and output representation into simpler and non-reducible components:

\paragraph{Irrep Decomposition} Any orthogonal representation $\rho : G \rightarrow \mathbb{R}^{d_{\rho} \times d_{\rho}}$ can be decomposed as a direct sum of mutually orthogonal \textit{irreducible representations (irreps)} $\rho(g) = Q^T (\bigoplus_{i \in I} \psi_i(g))Q$, where $\psi_i \in \hat{G}$ is an irrep, $\widehat{G}$ is the set of irreps of the group $G$, $I$ is an index set ranging over $\hat{G}$, and $Q$ is the change of basis. 
The kernel constraint solution for arbitrary $\rho_{\text{in}}$ and $\rho_{\text{out}}$ can be derived from the irreps decompositions of the two representations, as detailed by \citet{e2-equivariant}. A more comprehensive discussion on irrep decomposition is presented in Section \ref{sec:peter-weyl}.

In this work, we focus on convolutional networks operating on point clouds and 3D voxel data, and the compact group $SO(3)$. We employ \textit{dense convolutions} for 3D volumetric voxels, where filters are applied across all input data. For point clouds, we leverage message passing between neighboring points to effectively extract features from the unstructured data. 

\paragraph{Nonlinear Layers} In equivariant networks, nonlinear layers must also satisfy the equivariance constraint. While the transformation law of most feature field types do not commute with point-wise nonlinearities, pointwise nonlinearities acting on the norm of each field preserves their rotational equivariance.
This type of nonlinearity is called \textit{Norm nonlinearity}. Moreover, \textit{Gated nonlinearities}, which scales the norm of the feature fields using gated scalars, can be considered as a form of norm nonlinearity.
Lastly, \textit{Tensor product} nonlinearity combines feature vectors through tensor operations, which inherently introduces polynomial nonlinearities. Formal definitions of those nonlinear layers are provided in Sec. \ref{sec: other nonlinearities}.

\section{Regular and Quotient Nonlinearities}\label{sec: Regular and Quotient nonlinearities}

Although norm and gated nonlinearities satisfy the equivariance constraint, 
it has been observed that they typically under-perform compared to pointwise nonlinearities \cite{e2-equivariant}.
Unfortunately, while they are straightforward to implement for discrete groups, they require some form of discretization when considering continuous groups.
\citet{equivariant_mesh_cnn} first employed \textit{pointwise non-linearities} for the 2D rotation groups by leveraging a discretized Fourier transformation. 
Previously, \citet{cohen2018spherical} used a similar idea to implement group convolution networks over the 3D rotation group.
In this section, we delve in the theory behind this idea.

\paragraph{Peter-Weyl : Orthonormal Basis of Matrix Coefficients} According to the Peter-Weyl theorem (Thm. \ref{thm:orthonormal basis}), the matrix coefficients of (complex) irreps of a compact group \(G\), i.e.
\begin{align}
    \left\{\sqrt{d_\psi} \psi_{ij}(g) : G \to \mathbb{C} \ |\ \psi \in \widehat{G}, 1 \leq i, j \leq d_\psi \right\}
\end{align}
form a complete orthonormal basis for the vector space $L^2(G)$ of square-integrable functions over $G$, where $d_\psi$ is the dimensionality of the irrep $\psi$ and $\sqrt{d_\psi}$ is a scalar factor to normalize the basis.
This generalizes the classical \textit{Fourier transform} of periodic functions (corresponding to $G=SO(2)\cong U(1)$), where the notion of frequencies is replaced by the irreducible representations $\widehat{G}$ of the group $G$.
See Sec.~\ref{sec:peter-weyl} for more details.

\paragraph{Assumption}
In this work we are mostly interested in the group $G=SO(3)$ and real valued functions.
For this reason, from now, we will assume only real valued representations (and irreps) and functions.
The Peter-Weyl theorem above requires some minor adaptations in this case for certain groups, but it still holds exactly for $G=SO(3)$.
See \citet{escnn} for a precise statement of the theorem in this case and its implications for other groups.
Moreover, as common in the literature, we assume all representations to be orthogonal, satisfying $\rho(g^{-1}) = \rho(g)^{-1} = \rho(g)^T$.

\paragraph{Fourier transform}
For convenience, the Fourier coefficients of a function $f \in L^2(G)$ are typically aggregated by the irrep they belong to.
Hence, the Fourier transform $\hat{f}$ is matrix-valued:
\begin{align}
\label{eq:fourier_transform}
    \hat{f}(\psi) := \int_G f(g) \sqrt{d_\psi} \psi(g) dg \in \mathbb{R}^{d_\psi \times d_\psi}
\end{align}
for all $\psi \in \widehat{G}$.
Similarly, the inverse Fourier transform is defined as 
\begin{align}
\label{eq:inverse_fourier_transform}
f(g) = \sum_{\psi \in \hat{G}} \sqrt{d_\psi} Tr(\hat{f}(\psi)\psi(g)^T)
\end{align}
where $Tr(A B^T)$ is the standard matrix (Frobenius) inner product.

\paragraph{Regular representation}
The group $G$ carries an orthogonal action on $L^2(G)$ by translating functions via $g: f \mapsto g.f$, with $[g.f](h) := f(g^{-1}h)$.
This action is the \textit{regular representation} $\rho$ of the group; steerable CNNs employing regular representations as intermediate feature type are equivalent to group convolution networks, hence the popularity of this design choice.
Unfortunately, this construction is not practical when $G$ is a continuous group since the space $L^2(G)$ is infinite dimensional.
The Fourier transform just introduced provides an effective solution as \textit{bandlimited functions} in $L^2(G)$ can be represented by a finite number of Fourier coefficients, i.e. a finite subset $\tilde{G} \subset \widehat{G}$ of the group's irreps.

\paragraph{Quotient representation}
A similar construction exists when considering functions over a \textit{quotient space} $Q = G/H := \{gH | g \in G\}$ (with $H < G$ a subgroup) rather than $G$ itself.
Recall that an element of $Q$ is \textit{coset}, i.e. an equivalence class of elements of the form $gH = \{gh | h \in H\}$.
A function $f \in L^2(Q)$ is then equivalent to a function in $L^2(G)$ invariant under the right action of the subgroup $H$, i.e. $f(gh) = f(g)$ for all $h \in H$.
In particular, this subspace $L^2(Q) \subset L^2(G)$ is typically spanned by a smaller set of matrix coefficients than $L^2(G)$, providing a more compact representation than the regular representation.
A standard example is given by spherical signals for $Q = S^2 \cong SO(3) / SO(2)$, which we will also use later in this work.
Because the regular representation is a special case for $Q = G / \{e\} \cong G$, we will often use $Q$ to refer to the underlying space on which functions are defined and generally discuss quotient non-linearities. 
Moreover, since the quotient space $Q$ is not necessarily finite, the previous considerations about bandlimiting the regular representation apply for quotient representations too.

Finally, we note that this Fourier transform is precisely the change of basis matrix which performs the \textit{irreps decomposition} described in Sec.~\ref{sec:steerable cnns} of the regular or quotient representation.
Indeed, the Fourier coefficients jointly transform under a direct sum of irreps, in the same way each frequency component in the Fourier transform of a periodic function transforms independently when the function is translated.
This is an important aspect for our method, so we derive this decomposition explicitly in Sec.~\ref{sec:Inverse Fourier Transform}.



 \vspace{-.25cm}
\paragraph{Pointwise Non-Linearity via Fourier transform}
Representing an intermediate (bandlimited) regular or quotient type feature by its Fourier coefficients requires an additional step to incorporate pointwise non-linearities in the network.
This is done by composing the activation function $\sigma$ with a \textit{sampling step} (via a discretized inverse Fourier transform, or IFT) and a Fourier transform (FT):
\begin{align}
\label{eq:activation layer}
    f'(x) &= \left[\text{FT} \circ  \sigma \circ \text{IFT} (f) \right](x)
\end{align}
where FT recovers the bandlimited coefficients of the output function from a finite set of samples, while IFT specifically computes the value of the bandlimited function $f(x)$ at selected points in space $Q$.
It is important to note that this operation is only \textbf{approximately equivariant} and the degree of equivariance mostly relies on the number $N$ of samples used and their distribution over the group.
Uniform and well-spaced sampling in $Q$ is necessary to ensure accurate representation and generalization over the group transformations.
To create a \textit{sample set} $\Gamma \subset Q$, it is possible to use elements from a discrete subgroup of $G$, or alternatively distribute $N$ points within $Q$ by simulating a particles repulsion system as proposed by \citet{bspline} and implemented in \citet{escnn}.




\subsection{Irreps Decomposition of Bandlimited Quotient Representations}\label{sec:Inverse Fourier Transform}

In this section, we give an explicit construction of the irreps decomposition of a band-limited regular or quotient representation.
This is necessary for implementing them in the framework of steerable CNNs, since solving the kernel constraint relies on the irreps decomposition of the intermediate features.

As previously mentioned, for a quotient space $Q = G / H$, a function $f \in L^2(Q) \subset L^2(G)$ is equivalent to a function over $G$ invariant with respect to right action by $H$.
By considering the Fourier transform Eq.~\ref{eq:inverse_fourier_transform} of $f$, we find that
 \begin{align}
    f(gh) &= \sum_{\psi \in \hat{G}} \sqrt{d_\psi} Tr(\hat{f}(\psi) \psi(h)^T \psi(g)^T)
\end{align}
Hence, the function has the desired invariance only if $\hat{f}(\psi) \psi(h)^T = \hat{f}(\psi)$ for any $h \in H$.
Under a proper choice of basis\footnote{
    This invariance constraint identifies the $H$-invariant sub-representations of $\psi$.
    Indeed, the irrep $\psi$ can be thought as a representation of $H$ via \textit{restriction}.
    This representation is not necessarily irreducible anymore but can be further reduced in a direct sum of irreps of $H$. Check the discussion about induced representations in \citet{escnn} for more details.
    For simplicity, here we assume $\psi$ is already expressed in a basis such that its restriction to $H$ already has an irreps direct sum structure.
    In this way, the invariant components of $\hat{f}(\psi)$ corresponds to its columns which are associated with a trivial representation of $H$.
}
for $\psi \in \widehat{G}$, only certain columns of $\hat{f}(\psi)$ satisfy this invariance; 
then, let $P_\psi \in \mathbb{R}^{d_\psi \times q_\psi}$ be a mask matrix selecting only those $q_\psi$ columns of $\psi$ that are invariant.
For notational convenience, we also assume that $\hat{f}(\psi) \in \mathbb{R}^{d_\psi \times q_\psi}$ contains only these $q_\psi$ invariant columns (the other columns would contain only zero coefficients).
Then, we can express the function $f$ as:
\begin{align}
    f(gh) &= \sum_{\psi \in \hat{G}} \sqrt{d_\psi} Tr(\hat{f}(\psi)P_\psi^T\psi(gh)^T)\\
          &= \sum_{\psi \in \hat{G}} \sqrt{d_\psi} Tr(\hat{f}(\psi)P_\psi^T\psi(g)^T).
\end{align}
Next, using the following identities 
$Tr(AB^T) = \text{vec}(B)^T\text{vec}(A)$ 
and 
$\text{vec}(A B) = (\bigoplus_i^d A) \text{vec}(B)$, where $d$ is the number of columns of $B$ and $\bigoplus_i^d A$ is the direct sum of $d$ copies of $A$ (i.e. stacked along the diagonal),
we write:
\begin{align}
\label{eq:final_inverse_fourier}
    f(g) &= \sum_{\psi \in \hat{G}} \textcolor{green_}{\sqrt{d_\psi}}\left( \textcolor{blue_}{\left(\bigoplus^{q_\psi} \psi(g)\right)} \textcolor{green_}{vec(P_\psi)}\right)^{\textcolor{transpose_}{T}} \textcolor{red_}{vec(\hat{f}(\psi))}  \\
\label{eq:delta f}
    f(g) &= \textcolor{green_}{\hat{\delta}^T} \textcolor{blue_}{\bigoplus_{\psi \in \hat{G}}\bigoplus^{q_\psi} \psi(g)^T} \textcolor{red_}{\hat{f}} 
\end{align}
with $\textcolor{green_}{\hat{\delta} = \bigoplus_{\psi} \bigoplus^{q_\psi}\!\sqrt{d_\psi} vec(P_\psi)}$ 
and $\textcolor{red_}{\hat{f} = \bigoplus_{\psi}vec(\hat{f}(\psi))}$, where $\bigoplus$ concatenates vectors.
In this formulation, $\textcolor{green_}{\hat{\delta}}$ can be interpreted as a vector with the Fourier coefficients of an indicator function on the coset $eH \in Q$ - i.e. $\delta_H(g)$ is non-zero iff $g \in H$ - or just a Dirac delta centered on the origin $e\in G$ in case of a regular representation.
Eq.~\eqref{eq:delta f} can be simplified further to
\begin{align}\label{eq:inverse3}
    f(g_i) = \textcolor{green_}{\hat{\delta}^T } \textcolor{blue_}{\rho(g_i)^T} \textcolor{red_}{\hat{f}}
\end{align}
where  $\textcolor{blue_}{\rho(g_i)^T = \bigoplus_{\psi \in \hat{G}}\bigoplus^{q_\psi} \psi(g_i)^T}$.

In Eq. \eqref{eq:inverse3}, $\rho(g_i)$ represents the action of group element $g_i$ on the vector containing all Fourier coefficients $\hat{f}$.
When we use a finite subset of irreps $\Tilde{G} \subset \widehat{G}$, this vector is finite dimensional and we refer to $\rho$ as a \textit{band-limited} quotient (or regular) representation and we denote its size as $F = \sum_{\psi \in \Tilde{G}} d_\psi \cdot q_\psi$.


\subsection{Activation Layer}
\begin{figure}[!h]
    \centering
\includegraphics[width = 1\columnwidth]{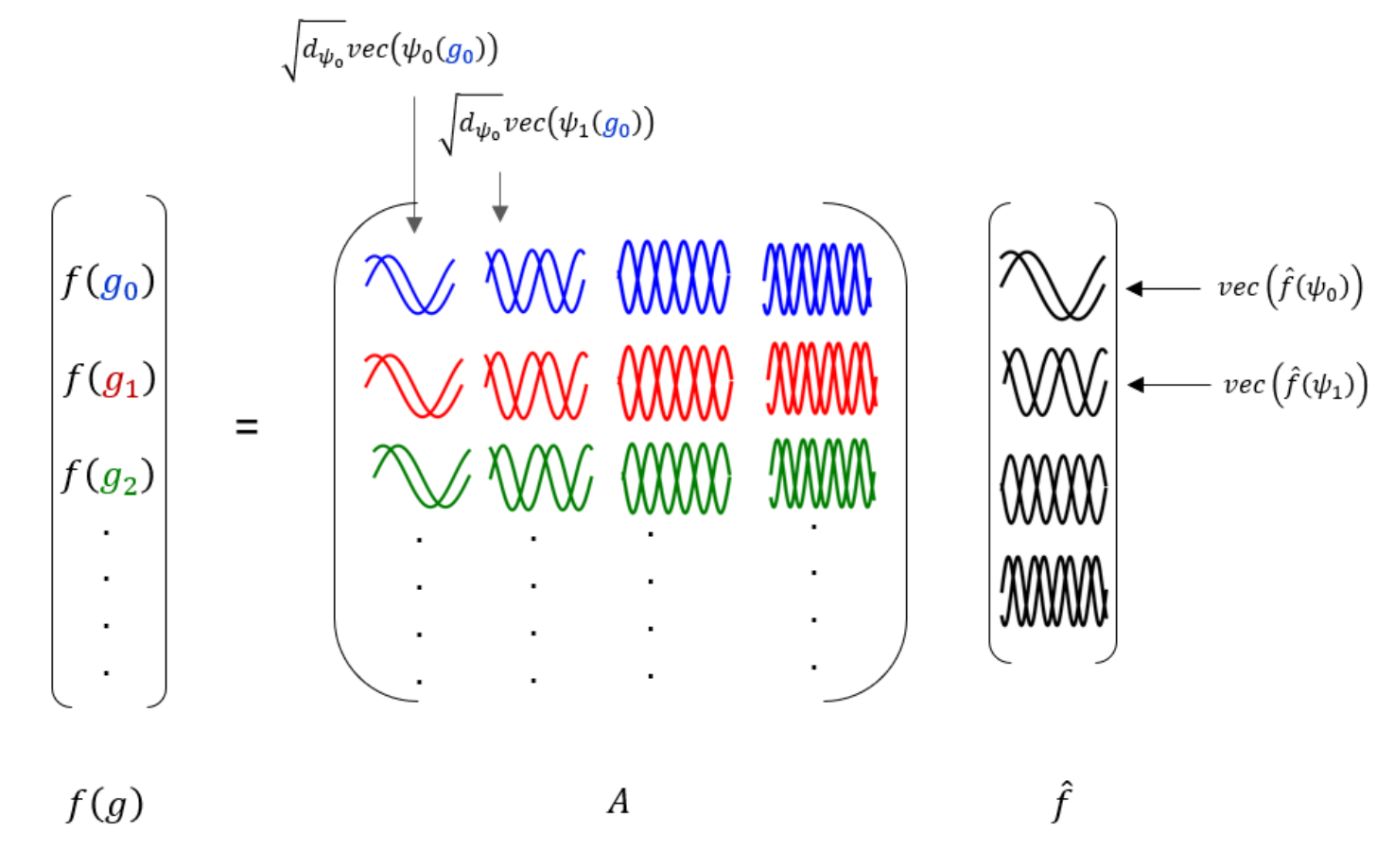}
    \caption{\footnotesize Discretized Inverse Fourier transform, where $\hat{f} = \bigoplus_\psi vec(\hat{f}(\psi)) \in \mathbb{R}^{F}$ and $A_i:= \rho(g_i)\hat{\delta} \in \mathbb{R}^F$. $N$ is the number of group samples, while $F$ is the size of the representation.
    Each row in $A$ corresponds to a single group sample, while each column corresponds to an irrep's matrix coefficients.}\label{fig:inverse_fourier}
\end{figure}

In this section we construct the activation layer described in Eq.~\ref{eq:activation layer} by leveraging a finite set of $N$ samples $\Gamma \subset Q$.
We first group the terms $\textcolor{blue_}{A_i :=\rho(g_i)\hat{\delta}}$ in Eq.~\eqref{eq:inverse3}:
\begin{align}
\label{eq:final inverse transform}
    f(g_i) &= \textcolor{blue_}{A_i^T}\textcolor{red_}{\hat{f}} 
\end{align}
Note that $\textcolor{blue_}{A_i :=\rho(g_i)\hat{\delta}}$ is the Fourier transform of an indicator function centered on the coset $g_i H \in Q$.
Hence, we define the matrix $A$ with a row $A_i$ for each $g_iH$ in the sampling set $\Gamma \subset Q$.
Conversely, the columns of $A$ correspond to the finite subset of irreps $\Tilde{G}$ used for band-limiting.
See Fig.~\ref{fig:inverse_fourier}. 
Essentially, the matrix $A$ carries out the discretized Inverse Fourier Transform (IFT) of $\hat{f}$; we refer to this matrix as the \textbf{sampling matrix}, as it integrates the group sampling.

Fourier and inverse Fourier transform are inverse operations; as $A$ implements the inverse Fourier transform, it suggests using its inverse to performing a discretized Fourier transform.
Because the activation function $\sigma$ introduces small high-frequency components in the function, we typically rely on oversampling to control \textit{aliasing}, i.e. $N > F$ and, therefore, we leverage the \textit{pseudo-inverse} matrix $A^{\dag}$, rather than the matrix inverse.
We provide more details on Fourier and inverse Fourier transforms in Apx.~\ref{sec:fourier and inverse fourier transform}.

The activation layer in Eq.~\ref{eq:activation layer} can then be formulated using the sampling matrix $A$ as
\begin{align}\label{eq:activation layer 2}
    \hat{f'} = A^\dag \sigma(A\hat{f}(x))
\end{align}
where $\hat{f}(x)$ is the input feature vector, and $\sigma$ is a pointwise nonlinearity such as ELU. 

Recall that this operation is only \textit{approximately equivariant} due to the \textit{aliasing} effect mentioned ealier.
To improve the equivariance of the nonlinearity layer, it's crucial to increase and uniformly distribute the number of group samples over the space $Q$; however, as explored in Sec.~\ref{sec:computational aspects}, the computational cost of this operation scales with the number $N$ of samples.

\subsection{Adaptive Sampling Matrix}\label{sec:adaptive sampling matrix}

In a typical architecture, a predefined number of samples is stored for repeated use across the network.
In this work, we dynamically learn the sampling matrix as an equivariant function $A(x)$ of the model's input $x$ to ensure full equivariance regardless of the number $N$ of samples:
\begin{align}
    A(g.x) = A(x) \rho(g)^T \ .
\end{align}
This strategy reduces the need for extensive sampling, effectively decreasing both the number of group samples \(N\) and the dimensions of \(A\), thereby can lower the overall computational complexity.
Additionally, we expect this approach to generate more expressive sampling matrices that can be specifically tailored to each input's salient directions.

In practice, we can generate the sampling matrix $A$ using an equivariant layer processing the model input or intermediate features, depending on the implementation choice. 
We explore different strategies in Sec.~\ref{sec:implementation}.

Unfortunately, computing the pseudoinverse (as well as the inverse) of a matrix can be computationally expensive and difficult to backpropagate through, especially for large matrices. 
However, if sufficiently many samples $N$ are used, the Peter-Weyl theorem ensures the columns of $A$ become orthogonal to each other.
Assuming the columns are sufficiently orthogonal, we can then approximate the pseudoinverse as $A(x)^{\dag} \approx \frac{1}{N} A(x)^T$, further improving the computational efficiency.
The final layer takes the form 
\begin{align}\label{eq:final activation layer}
    \tilde{f} = \frac{1}{N}A(x)^T\sigma(A(x)\hat{f}(x)).
\end{align}
If the sampling matrix is generated by an equivariant layer, the nonlinear layer in Eq.~\ref{eq:final activation layer} is fully equivariant regardless of the number $N$ of samples considered.
We prove this result in Apx.~\ref{sec: Equivariance of the nonlinear layer}.

\begin{figure*}[htbp] 
\begin{tikzpicture}[
  node distance=0.5cm and 1cm,
  mynode/.style={
    draw,
    rounded corners,
    align=center,
    fill=#1,
    minimum height=1cm,
    minimum width=1.5cm,
    font=\scriptsize,
    inner sep=2pt 
  },
  myarrow/.style={-Stealth, thick, draw = mycustomcolordark},
  arrowlabel/.style={midway, fill=white, above, font=\small}, 
  rightnode/.style={
    draw,
    rounded corners,
    align=center,
    fill=#1,
    minimum height=1cm,
    minimum width=1.5cm,
    font=\tiny
  },
  midnode/.style={ 
    draw,
    rounded corners,
    align=center,
    fill=silver!60, 
    font=\scriptsize,
    above=6mm, 
    minimum height = 1cm,
    minimum height = 1cm
  }
]
\node[mynode=mycolor] (conv1) {Conv.\\+\\Norm.};
\node[mynode=mycolor!60, right=of conv1] (block2) {Nonlinear\\layer}; 
\node[mynode=mycolor, right=of block2] (block3) {Conv.\\+\\Norm.};
\node[mynode=mycolor!60, right=of block3] (block4) {Nonlinear\\layer}; 
\node[mynode=mycolor, right=of block4] (block5) {Conv.\\+\\Norm.};
\node[mynode=mycolor!60, right=of block5] (block6) {Nonlinear\\layer}; 
\node[mynode=mycolor, right=of block6] (block7) {Pooling};

\draw[myarrow] (conv1) -- node[arrowlabel] (fhat) {$\hat{f}$} (block2);
\node[midnode] at (fhat.north) (midpoint1) {Sampling \\ branch}; 
\node[midnode, right=of midpoint1, xshift = 11mm] (midpoint2) {Spatial \\ downsampling}; 
\node[midnode, right=of midpoint2, xshift = 18mm] (midpoint3) {Spatial \\ downsampling}; 

\draw[myarrow] (fhat) -- (midpoint1.south);
\draw[myarrow] (midpoint1) -- node[arrowlabel] {$A$} (midpoint2);
\draw[myarrow] (midpoint2) -- node[arrowlabel] {$A$} (midpoint3);
\draw[myarrow] (midpoint3.east) -- ++ (2.5,0) node[midway, above] {$A$};; 

\draw[myarrow] ($(midpoint1)!0.35!(midpoint2)$) to ([xshift=-0.2mm]block2.north);

\draw[myarrow] ($(midpoint2)!0.6!(midpoint3)$) to ([xshift=0mm]block4.north);

\coordinate (finalpoint) at ([xshift=2.42cm]midpoint3.east);
\draw[myarrow] (finalpoint) -- (block6.north);

\coordinate (input) at ([yshift=2cm]conv1.north);

\draw[myarrow] (input) -- (conv1);

\draw[myarrow] (conv1) -- node[arrowlabel] {$\hat{f}$} (block2);
\draw[myarrow] (block2) -- node[arrowlabel] {$\hat{f}$} (block3);
\draw[myarrow] (block3) -- node[arrowlabel] {$\hat{f}$} (block4);
\draw[myarrow] (block4) -- node[arrowlabel] {$\hat{f}$} (block5);
\draw[myarrow] (block5) -- node[arrowlabel] {$\hat{f}$} (block6);
\draw[myarrow] (block6) -- node[arrowlabel] {$\hat{f}$} (block7);
\draw[myarrow] (block7) -- ++(1.25,0); 

\node[inner sep=0pt] (input) at (0,3.2) 
          {\includegraphics[width=2cm]{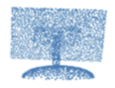}};
\end{tikzpicture} 
\caption{\footnotesize Architecture for point cloud processing. In this architecture, blue blocks at the bottom represents the main branch which process the point clouds and the gray blocks at the top correspond to sampling branch which generates the sampling matrix $A$ and perform spatial downsampling accordingly. Although it is not illustrated in the figure, each convolutional block, which comprises the convolutional layer, batch normalization and the nonlinear layer, is followed by an equivariant MLP.   }
    \label{fig:architecture_modelnet}
\end{figure*}
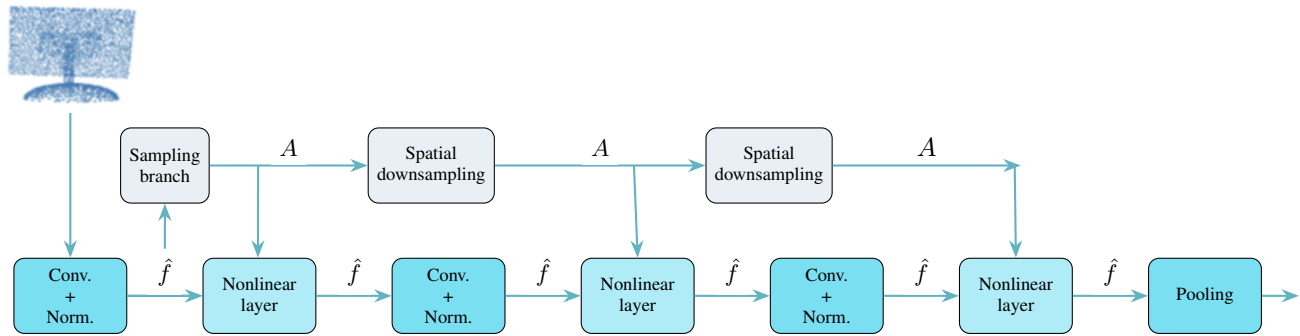

\section{Computational aspects}\label{sec:computational aspects}

Assume an input feature field $\hat{f}$ comprising $c$  $\rho$-fields, i.e. \(\hat{f}(x) = \bigoplus_i^c f_i(x) \in \mathbb{R}^{cF}\), where $\rho$ is a quotient (or regular) representation of size $F$.
Accordingly, the computational complexity for the inverse Fourier transform \(A\hat{f}\) and the Fourier transform \(A^{\dag}\sigma(A\hat{f})\) is \(O (cNF)\), with \(N\) representing the number of group samples. 
The pointwise nonlinearity \(\sigma\) has a complexity of \(O(cN)\).
For a data sample containing \(m\) pixels / spatial samples, this yields a total complexity for the nonlinear layer of \(O(mcNF) + O(mcN) + O(mcFN) = O(mcNF)\), indicating that the layer's computational complexity scales with the number of group samples \(N\).

In the context of the convolutional layer, let \(k\) denote the computational cost for a single convolution operation involving one input and one output channel, while \(m\) represents the number of pixels / spatial samples in the input grid. 
For a convolutional layer with $c$ input and output channels, the complexity becomes \(O(km c^2 F^2)\).
Then, the model has overall complexity \(O(km c^2 F^2) + O(mcNF)\). 
Whenever $kcF >> N$, the complexity is dominated by the convolution term \(O(km c^2 F^2)\), making the impact of the number of samples $N$ less relevant, so we can not expect strong improvements in the final runtime.
However, because of oversampling, $N>F$ and, therefore, the intermediate activations of the non-linear layers are larger than the features processed by the convolution layers and can increase the overall memory requirements of the model.
We study the effect of oversampling and of adopting our method on memory usage in our experiments.
Finally, in Sec.~\ref{sec:conclusion} we discuss potential future research directions to employ our method and benefit more from its reduced sampling rate.

\section{Implementation}\label{sec:implementation}
We can employ two main approaches to generate the sampling matrix $A$. Either a unique set of sampling matrices can be generated for each nonlinear layer in the network, or a set of matrices can be generated only once, and it is processed by all following nonlinear layers. In our implementation, we focused on the latter approach to obtain more expressive $A$ and to increase computational efficiency. Empirical results also showed that the latter approach indeed outperforms the former in terms of efficiency. 

 Our approach generates a unique sampling matrix $A$ for each point in the point cloud or for each pixel in 3D voxels. In both cases, the main architecture employs some form of spatial downsampling.
 As the data is progressively downsampled through the layers, a compatible downsampling technique should also be applied to the sampling matrices to ensure their spatial resolution remains compatible with the feature space.
 Regarding the point clouds, this downsampling can be executed by simply indexing - i.e. retaining only the sampling matrices of the points that remain after downsampling. Alternatively, a convolutional layer can be used for a smoother downsampling for both data types.  

Furthermore, the sampling matrix can be generated using either an equivariant MLP or an equivariant convolutional layer. The main advantage of employing a convolutional layer is its ability to utilize neighbouring information. While downsampling through convolutional layers is often needed for matrices generated by an equivariant MLP, it is not required for matrices produced by equivariant convolutional layers. Fig. \ref{fig:architecture_modelnet} illustrates the architecture for the point cloud processing. The blue blocks at the bottom represent the main branch in the network, while the gray blocks at the top illustrates our approach, where the matrices are generated by processing the intermediate feature and downsampled through the network.  In Sec.~\ref{sec:Experiments}, we present the results from the selected architectures that demonstrate the best performance among various architectures. We also discuss the practical design choices in Apx.~\ref{sec:practical design choices}.

\section{Related Work}

\paragraph{Equivariant Neural Networks} The concept of group equivariance in convolutional networks was first introduced by  \citet{group_CNN}, showing how CNNs can adapt to transformations like rotations and reflections. However, scaling to larger groups introduces computational challenges due to the need for extensive sampling. Steerable CNNs, also developed by \citet{steerable_cnn}, address the computational challenges by using the steerable kernels, which improves the flexibility of the network and reduces the computational overhead. Further exploration of steerable CNNs has been conducted in the 2D domain \cite{steerable_cnn,e2-equivariant, harmonic_networks, learning_steerable_filters}, and their expansion into the 3D domain \cite{3D_steerable_cnns, tensor_field_networks, 5dbb864614c2401086812e6f25ec26f7, escnn} has demonstrated significant benefits in processing point clouds and graphs \cite{tensor_field_networks,covariant_molecular_networks,equivariant_pointnet++}. 

\paragraph{Equivariant Nonlinearities}
One of the challenges in designing rotation equivariant networks is selecting appropriate activation functions. To maintain equivariance, nonlinearities must commute with the rotation of equivariant features, which cannot be achieved by conventional pointwise nonlinearities. To address this, norm and gated nonlinearities have been introduced by \citet{tensor_field_networks} and \citet{3D_steerable_cnns}, respectively, which apply to the norm of equivariant features and preserve rotation commutativity, but cannot capture directional information. An alternative approach, as proposed in \citet{covariant_molecular_networks, Clebsch_Gordan_Nets}, involves using tensor product operations on equivariant features as a form of nonlinearity, which is further discussed in \citet{n_body_networks}. 

\citet{equivariant_mesh_cnn} introduce a nonlinearity built on Fourier transformation, where they interpret the features as Fourier coefficients of functions over the circle, which we also discussed in Sec.~\ref{sec: Regular and Quotient nonlinearities}. In a concurrent work, \citet{general_nonlinearities_on_so2_equivariant_cnns} propose a Fast Fourier Transform based approach to apply conventional pointwise nonlinearities on equivariant representations and obtain exact $SO(2)$ equivariance for polynomial functions.

\paragraph{Adaptive Sampling and Canonicalization for Equivariant Networks}

In conventional CNNs, convolutional operations are accompanied by downsampling methods such as pooling and strided convolutions. The fixed grid structure in downsampling methods can compromise the network's ability to handle shifted and rotated inputs, thereby disrupts the shift invariance and equivariance \cite{shift_invariance_1, shift_invariance_2}. To address these limitations, \citet{group_equivariant_subsampling} introduced a new subsampling method that uses input-dependent grids instead of fixed ones, maintaining translation equivariance. Additionally, \citet{adaptive_polyphase_sampling_APS}  proposed adaptive polyphase sampling (APS) to dynamically select downsampling grids.  Building upon this,  \citet{learnable_polyphase_sampling} improves APS by making this selection process learnable, using neural networks. \citet{equivariance_with_learnable_canonicalization_functions} took a different approach by using equivariant networks to predict input poses, allowing the network to revert inputs to a rotation invariant state for processing in non-equivariant models. \citet{learning_probabilistic_symmetrization} expanded on this by predicting a distribution over possible poses, rather than a single pose, and averaging the model's output over samples taken from this distribution.
\citet{frame_averaging} introduce Minimal Frame Averaging (MFA), which can be thought as a deterministic version of the previous approach with a minimal number of samples; this strategy allows for exact equivariance and extends to a broad range of symmetry groups, including Lorentz and unitary groups.  
Finally, \citet{CRIN} take a local approach constructing local reference frames for each point in a point-cloud using the centrifugal reference frames (CRFs).
Similarly, \citet{sangalli2023moving} constructs local reference frames, by leveraging the eigenvectors of the input signal's Hessian at each spatial location.


In our study, we advance the concept of adaptive grids, previously applied in downsampling and canonicalization, by integrating it into Fourier-based nonlinearities, more specifically to the sampling process. Unlike the previous works that largely focused on utilizing global grids for downsampling or canonicalization, our approach takes a more localized perspective, learning a sampling grid at each spatial location.

\section{Experiments}\label{sec:Experiments}
Methods that are explained in Sec \ref{sec:adaptive sampling matrix} and in Sec. \ref{sec:implementation} are evaluated on two datasets, namely ModelNet10 \cite{modelnet} and NoduleMNIST3D \cite{medmnist}. ModelNet10 \cite{modelnet} is a subset of the larger ModelNet40 which contains synthetic object point clouds, while NoduleMNIST3D \cite{medmnist} is a subset of the MedMNIST collection. All our implementations are based on the \textit{escnn} library \cite{escnn}.

\begin{figure}[!h]
    \centering
    \includegraphics[width=\columnwidth]{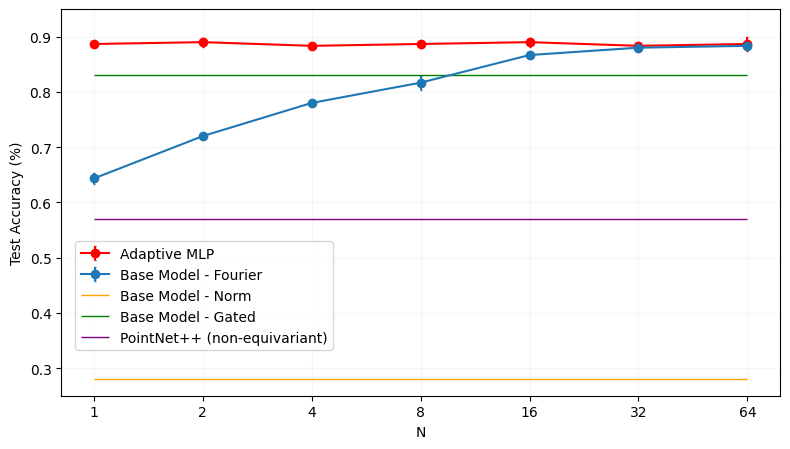}
    \vspace{-.75cm}
    \caption{\small Test accuracy on ModelNet10, with respect to the number of group samples by model. }
    \label{fig:modelnet_accuracies}
\end{figure}

\paragraph{ModelNet10}
In our analysis, we benchmarked our method against the approach described in \citet{equivariant_pointnet++}, which combines \textit{Tensor Field Networks (TFN)} \cite{tensor_field_networks} with Fourier-based nonlinearities. In this study, we developed a similar model consisting of three convolutional blocks, each followed by a block of linear layers. We refer to this model as the \textit{Base model}. Fig. \ref{fig:modelnet_accuracies} presents the test accuracies of the Base model with various nonlinearities. The figure demonstrates that the Base model with Fourier-based nonlinearities outperforms the models with other nonlinearities, such as norm and gated, provided a sufficient number of group elements are sampled. Additionally, the model with gated nonlinearity significantly outperforms the model with norm nonlinearity. This outcome is expected, since the gated nonlinearities are more proficient at processing directional information than norm nonlinearities. 
Furthermore, we have confirmed that the traditional PointNet{\footnotesize++} framework lacks the capacity to handle equivariance adequately.

In our adaptive approach, we experimented with various downsampling and matrix generation methods, while keeping the main architecture consistent. We only included the best-performing model, which is referred to as the \textit{Adaptive MLP}. Its architecture is depicted in Fig. \ref{fig:architecture_modelnet}. This model generates the sampling matrix through an equivariant MLP and performs spatial downsampling using convolutional layers. 
Fig \ref{fig:modelnet_accuracies} displays the performance of the models with different nonlinear layers,  based on the number of group samples. Each model is trained three times, with three different predetermined seeds for initialization. Although the standard deviations among the runs are quite small, they are visualized in the figure.  As shown, even with a single group sample, our Adaptive MLP outperforms the Base model. This improvement is attributed to its complete equivariance and enhanced expressiveness, achieved by introducing a tailored sampling matrix to each nonlinear layer. 

\begin{figure}[!h]
    \centering
    \includegraphics[width=\columnwidth]{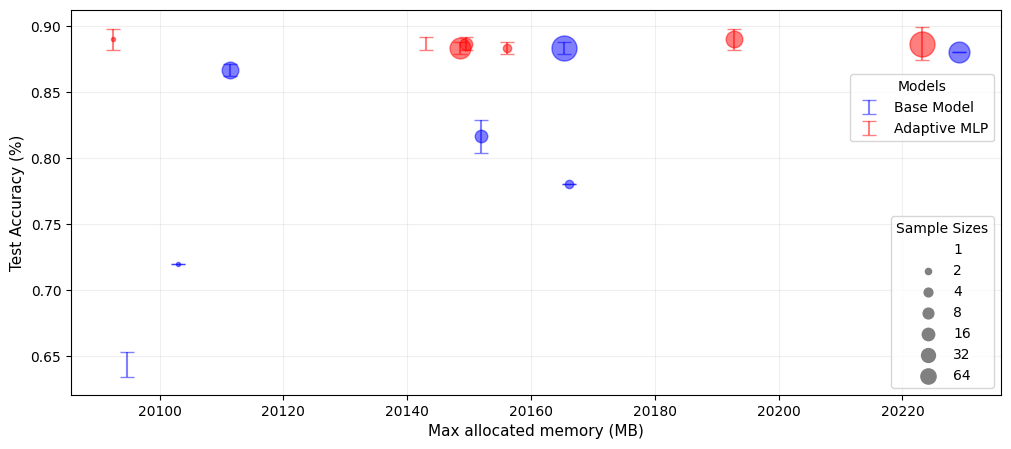}
    \vspace{-.75cm}
    \caption{\small Test accuracies vs. memory cost by model. Results are computed on ModelNet10.  }
    \label{fig:scatter_modelnet}
\end{figure}

We can also observe the memory gain obtained by our approach in Fig. \ref{fig:scatter_modelnet}. Our Adaptive MLP with two group samples, located in the top left corner of the figure, outperforms other models in both test accuracy and computational efficiency. However, it is important to note that the computational gains are modest and do not increase proportionally with the number of group samples, as initially expected. Further analysis reveals that this can be attributed to the computational overhead in the convolutional layer, which diminishes the improvements gained from the nonlinear layer, making them less apparent in the overall model output. We discuss the computational limitations in more detail in Sec. \ref{sec:computational aspects}.

\paragraph{NoduleMNIST3D} We employ an equivariant architecture with dense convolutions and Fourier-based nonlinearities as a benchmark to evaluate the performance of our approach. This model will be referred to as the \textit{Base model}. For the adaptive sampling approach, we have tried various implementations. Specifically, we examined three different methods for spatial downsampling of the sampling matrix.

Note that each model, with a specific number of group samples, is trained three times using different predetermined initialization seeds. In the first approach, convolutional layers are employed to perform all downsampling, and this model is referred to as \textit{Adaptive Conv}.
To further reduce the computational cost, in our second implementation, we replaced the convolutional layers performing downsampling with average pooling layers. This model is referred to as \textit{Adaptive Pooling}. However, both \textit{Adaptive Conv} and \textit{Adaptive Pooling} were outperformed by the \textit{Base model} in terms of test accuracy and computational efficiency. Our further analysis revealed that the memory overhead was primarily caused by the first convolutional layer in the sampling branch.

\begin{figure}[!h]
  \centering
    \vspace{-.15cm}  \includegraphics[width=\columnwidth]{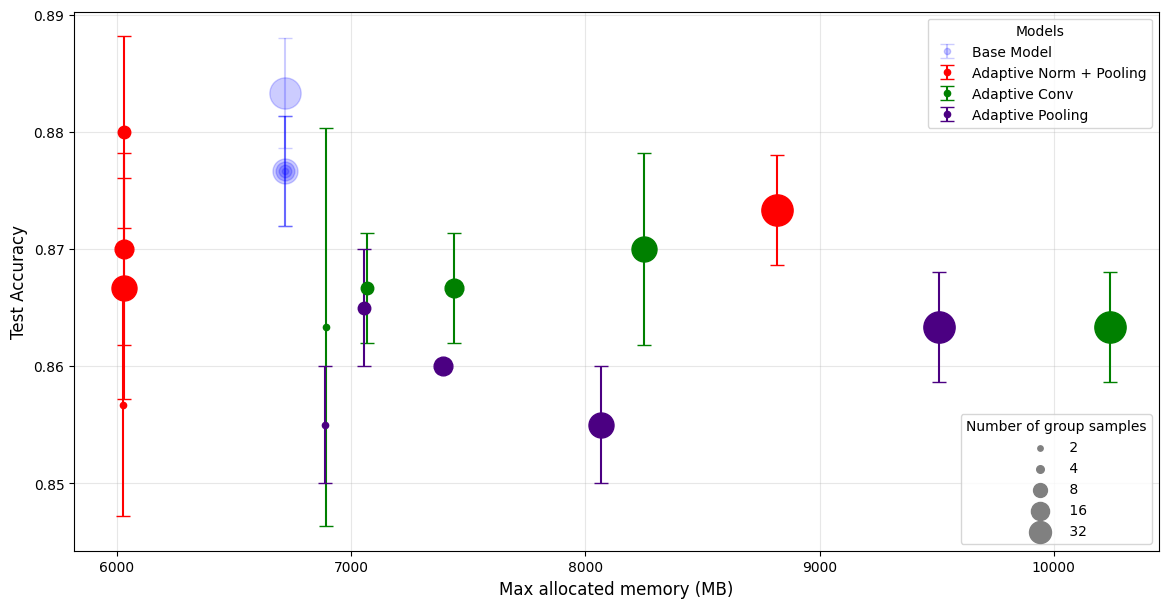}
  \vspace{-.75cm}
  \caption{\small Test accuracy vs. memory cost by model. The results are computed on NoduleMNIST3D datastet. Each model is trained three times with three different predetermined seeds for initialization. Standard deviation among those runs are displayed in the figure. The Base model is illustrated with transparency to display the overlap in the plot.  }
\label{fig:scatter_mnist_1}
\end{figure}

 This layer processes the initial intermediate features from the main branch and returns the first set of generated sampling matrices to the first activation layer (see Fig. \ref{fig:architecture_mnist}). To address this issue, we replaced the Fourier-based nonlinearity in the first convolutional block with the norm nonlinearity, and started generating the sampling matrix from the second convolutional layer onwards. This implementation is referred to as \textit{Adaptive Norm + Pooling} in the figures. The \textit{Adaptive Norm + Pooling} model successfully resolved the convolutional overhead issue, leading to improved computational efficiency.

In Fig. \ref{fig:scatter_mnist_1}, we present the test accuracy and the memory consumption for the different implementations. \textit{Adaptive Conv} and \textit{Adaptive Pooling}, represented by green and indigo points respectively, show lower test accuracies and higher computational demands compared to the base model as seen mainly on the right side of the figure. Conversely, Adaptive Norm + Pooling (red points) successfully reduces computational demands while achieving comparable test accuracies to the base model. It is important to note that the standard deviation values for the Adaptive Norm + Pooling with a smaller number of group samples are notably higher compared to the other models, which suggests more variability in their performance.

\section{Discussion and Conclusion}
\label{sec:conclusion}
In the previous sections, we presented the findings of our empirical analysis. When applied to point cloud data, our approach demonstrated improved classification accuracy compared to models that did not incorporate our method, demonstrating the benefits of maintaining exact equivariance and leveraging a local adaptive sampling grid. We also observed a marginal improvement in computational cost. However, when applied to voxel data, which lack exact symmetries due to discretization, our approach achieved moderate computational efficiency and comparable classification accuracy.

To better understand the limitations in computational efficiency, we analyzed the computational costs of each network layer using profiling tools. Our findings reveal that convolutional layers have a significantly higher computational load compared to nonlinear layers, particularly in point cloud processing. The cost ratio between convolutional and nonlinear layers varies throughout the network, with convolutional layers being up to seven times more computationally demanding than nonlinear layers. This substantial difference dominates the overall cost of the model, making the improvements in the nonlinear layers less noticeable in the overall computational cost of the model.

\paragraph{Further Improvements and Future Work}
To address the computational overhead in the convolutional layers, future works could adopt different model designs which aim to reduce the memory and computational overhead of convolutional layers, making the benefits of our method more observable.
For instance,  MobileNetV2 \cite{mobilenetv2} uses separable depthwise convolutions to significantly reduce computational load and could highlight the impact of nonlinear layers.
Similarly, implementing the full model from \citet{equivariant_pointnet++}, which uses entire MLPs as pointwise activations, can introduce higher complexity but benefit from a small and adaptive grid to manage computational cost and reduce aliasing effects. 
Finally, another promising alternative is the architecture from \citet{knigge2022exploiting}, which leverages a separable group convolution.

\section*{Acknowledgement}
We express our gratitude to the anonymous reviewers and Dr. Erik Bekkers for their invaluable feedback and insightful discussions.

\bibliography{example_paper}
\bibliographystyle{icml2024}


\null

\appendix

\onecolumn

\tableofcontents

\newpage

\section{Mathematical Background}

\subsection{Peter-Weyl Theorem}\label{sec:peter-weyl}
In the context of harmonic analysis, Fourier transform decomposes a function into its frequency components. The Peter-Weyl theorem can be considered as a generalization of the classical Fourier transform to the setting of compact topological groups. Similar to the Fourier transform, the Peter-Weyl theorem decomposes certain functions on a compact group $G$ into a series of irreducible components.

\begin{theorem}[thm:decomposition into irreps]{Peter-Weyl Theorem : Decomposition into orthogonal subspaces}
Given a compact group $G$, the space of square-integrable functions on $G$, which is denoted by $L^2(G)$,  can be expressed as a \textit{direct sum} of mutually orthogonal subspaces $V_{\pi}$;
\begin{align}
    L^2(G) = \bigoplus_{\pi \in \hat{G}} \bigoplus_{i=1}^{m_j} V_{\pi}
\end{align}
where $\hat{G}$ denotes the set of all equivalence classes of finite dimensional irreducible unitary representations of $G$, and $m_j$ is the multiplicities of the corresponding subspace. Each subspace $V_{\pi}$ is isomorphic to the representation $\pi$, and invariant under left and right translation by $g \in G$.
\end{theorem}

Note that this theorem only applies to the compact groups. 

Intuitively, groups can be interpreted as rotations acting on geometric objects. In the context of \textbf{compact groups} \footnote{A compact group is a group that is also a compact topological space, meaning it is both closed and bounded. Compact groups can be infinite in terms of the number of elements, but have a finite size from a topological perspective. $SO(3)$ is an example of compact groups.}, the matrix coefficients of their representations can be interpreted as square-integrable functions defined on the group. This perspective allows us to specifically redefine the decomposition of a unitary representation in terms of the group's representations.

\vspace{-2cm}
\begin{definition}[def:decomposition into irreps]{Decomposition into irreducible representations}
Given $\rho : G \rightarrow GL(V) $ is a unitary representation of a compact group $G$ over a field with characteristic zero, $\rho$ can be expressed as a direct sum of mutually orthogonal \textbf{irreducible representations} :

\begin{align}
        \rho(g) = Q^T \Bigg( \bigoplus_{i \in I} \psi_i(g) \Bigg) Q
\end{align}
where $\psi_i \in \hat{G}$ is an irrep, $\widehat{G}$ is the set of irreps of the group $G$, $I$ is an index set ranging over $\hat{G}$, and $Q \in GL(V)$ is the change of basis. 
\end{definition}

Note that Peter-Weyl theorem ensures that each irrep acts on an invariant subspace of $V$.

\begin{theorem}[thm:orthonormal basis]{Peter-Weyl : Orthonormal Basis}
Let $G$ be a compact group, $\rho : G \rightarrow GL(V)$ a unitary representation, and $L^2(G)$ vector space of square-integrable functions on $G$. Unitary representation $\rho$ can be decomposed into irreps, $\psi$, as shown in Def. \ref{def:decomposition into irreps}. Peter-Weyl Theorem guarantees that there are distinct linear subspaces of $L^2(G)$, each corresponding to an irreducible representation that transforms accordingly. \\

The matrix coefficients of the irrep $\psi$, denoted by $\psi_{ij}(g)$, form an orthonormal basis for each subspace $V_{\psi}$, and they satisfy the following orthonormality relations.\\

For infinite groups:
$$\int_G \overline{\psi'_{ij}(g)} \psi_{kl}(g)  \, dg = \frac{1}{d_{\psi}} \delta_{\psi\psi'} \delta_{ik} \delta_{jl} ,$$

and for finite groups:
$$ \frac{1}{|G|} \sum_{g \in G}  \overline{\psi'_{ij}(g)} \psi_{kl}(g) = \frac{1}{d_{\psi}} \delta_{\psi\psi'} \delta_{ik} \delta_{jl} $$

where $d_{\psi}$ is the dimension of the irrep $\psi$.\\

If the vector space $V$ is a complex space, the orthonormality condition implies that the matrix coefficients form a complete orthonormal basis $ \left\{\sqrt{d_\psi} \psi_{ij}(g) : G \to \mathbb{C} \ |\ \psi \in \widehat{G}, 1 \leq i, j \leq d_\psi \right\}$ for complex square integrable functions in $L^2(G)$.
\end{theorem}

In the case where the vector space is real $V = \mathbb{R}^n$, conjugated representations $\overline{\psi_{ij}^{l}(g)}$ is replaced by $\psi^l_{ij}$. 
From now on, we will assume vector spaces are of real type unless specified otherwise. It is important to note that for real irreps, some coefficients may be redundant. However, since we are exclusively working with $SO(3)$ in this study, this does not apply. For more details on this setting, please refer to \citet{escnn}.

Overall, the Peter-Weyl theorem provides a method for parameterizing functions over a group, which is especially useful for infinite groups.

\subsection{Fourier and Inverse Fourier Transform}\label{sec:fourier and inverse fourier transform}

In the context of the Peter-Weyl theorem (Thm.~\ref{thm:orthonormal basis}), any square-integrable function on a compact group \(G\), denoted \(f: G \rightarrow \mathbb{R}\), can be expanded as a series using the matrix coefficients of the group's irreducible representations.

\begin{align}\label{eq:first_inverse_fourier}
    f(g) = \sum_{l=0}^{L} \sum_{\psi^l \in \hat{G}} \sum_{m,n <d_{\psi}} w_{lmn} \cdot \sqrt{d_{\psi^l}} \hspace{.1cm} \psi_{mn}^l(g)
\end{align}
where $l$ is the degree of the spherical harmonics, $\hat{G}$ is the set of all irreducible representations of $G$, $d_{\psi}$ is the dimension of irrep $\psi : G \rightarrow \mathbb{R}$ and the $\psi^l_{mn}(g)$ are the matrix coefficients of $\psi$. Coefficients $w_{lmn}$ parameterize the function 
$f$ on this basis, while $\sqrt{d_{\psi}}$ ensures that the basis is normalized.

This expansion is analogous to a Fourier series, where sines and cosines are replaced by the matrix coefficients of irreps. In this setting, coefficients $w_{lmn}$ serve as the Fourier coefficients, which can be formulated as
$$ w_{lmn} = \int_G f(g)\psi^l_{mn}(g)dg .$$

where $dg$ is the normalized Haar measure on the group $G$. Haar measure can briefly be defined as the prior probabilities for compact groups of transformations, and it allows for the integration of functions over the group. \\
The projection onto the full set of entries of irrep $\psi$ can be reformulated  and referred to as the \textbf{Fourier transform}, as in Def \ref{def:fourier transform}.

\begin{definition}[def:fourier transform]{Fourier Transform}
    For the compact group $G$, let $f : G \rightarrow \mathbb{R}$ be the square-integrable functions on $G$, and $\psi :G \rightarrow \mathbb{R}$ be the irreps of $G$. \textbf{Fourier transform} over the function $f$ is formulated as follows for
    \begin{align}
       \text{infinite groups:} \quad \hat{f}(\psi) &= \int_G f(g) \sqrt{d_\psi} \psi(g) dg \quad \in \mathbb{R}^{d_\psi \times d_\psi} \\
       \text{} \nonumber \\
       \text{\noindent finite groups:} \hspace{10pt} \quad  \hat{f}(\psi) &= \frac{1}{|G|} \sum_{g \in G} f(g) \sqrt{d_{\psi}} \psi(g) \quad \in \mathbb{R}^{d_\psi \times d_\psi} \label{eq:FT}
    \end{align}
where $\hat{f}(\psi)$ is the Fourier coefficients of the corresponding irrep $\psi$ and $\sqrt{d_{\psi}}$ is the dimension of the irrep $\psi$.
\end{definition}

It is important to note that the Fourier transform exhibits the equivariance property with respect to the action of a group element \(g \in G\) on a function \(f : G \rightarrow \mathbb{R}\):
\[
\widehat{g \cdot f}(\psi) = \psi(g)\widehat{f}(\psi)
\]
for any irreducible representation (irrep) $\psi$.

Regarding the expansion of the square-integrable functions on $G$, the function in Eq. \eqref{eq:first_inverse_fourier} can be redefined in a more compact form with the help of trace operation, which is defined as:
\begin{align}
    Tr(A^TB) = \sum_{m,n <d} A_{mn}B_{mn} \in \mathbb{R}
\end{align}
where $A,B \in \mathbb{R}^{d \times d}$. Since coefficients $w^l_{mn}$ in Eq. \eqref{eq:first_inverse_fourier} serves as Fourier coefficients, $\hat{f}(\psi^l) \in \mathbb{R}^{d_\psi \times d_\psi}$ can be defined as the matrix containing the coefficients $w^l_{mn} \in \mathbb{R}$. Given that, the \textbf{Inverse Fourier Transform} is defined.

\begin{definition}[def:inverse fourier transform]{Inverse Fourier Transform}
Given group $G$ and a function on $G$, $f : G \rightarrow \mathbb{R}$, \textbf{Inverse Fourier Transform} can be defined as
      \begin{align}\label{eq:inverse fourier transform}
           f(g)= \sum_{\psi \in \hat{G}} \sqrt{d_{\psi}} \hspace{.11cm} Tr(\psi(g)^{T} \hspace{.02cm} \hat{f}(\psi)) 
      \end{align}
where $\hat{G}$ is the set of irreps of the group $G$, $\hat{f}(\psi)$ is the Fourier coefficients of irrep $\psi$, and $d_{\psi}$ is the dimension of the corresponding irrep.   
\end{definition}

For the infinite groups, there exists infinite irreducible representations. We can still parameterize a function using the same expression for the Inverse Fourier Transform (Def. \ref{def:inverse fourier transform}), but only by taking the finite subset of the irreps,  which is $\hat{G} \subset \Tilde{G}$, into account in the computation. This finite subset is also referred to as \textit{bandlimited} representations.

\subsection{From Fourier Transform to the Regular Representation}
The Peter-Weyl theorem and Fourier transform has a strong connection in terms of breaking down functions into their irreducible components. In this section, this connection will be shown in a more formal sense, focusing on the decomposition of the regular representation.

The inverse Fourier transform (Eq. \eqref{eq:inverse fourier transform}) is defined in the previous section. Transpose of an inner product of two matrices can be defined in terms of vectorization of the matrices, as in
$Tr(A^TB) = \text{vec}(A)^T\text{vec}(B).$ Given this, the trace in the inverse Fourier transform (Def. \ref{def:inverse fourier transform}) can be rewritten as
\begin{align}
    Tr(\psi(g_i)^T\hat{f}(\psi)) = \text{vec}(\psi(g_i))^T\text{vec}(\hat{f}(\psi)).
\end{align}

This makes the inverse Fourier transform (Eq. \eqref{eq:inverse fourier transform})
\begin{align}
    f(g) = \sum_{\psi \in \hat{G}} \sqrt{d_{\psi}} \text{vec}(\psi(g))^T \text{vec}(\hat{f}(\psi)
\end{align}

The summation can be converted into the matrix notation with the help of direct sum. Fourier coefficients $\hat{f}(\psi) \in \mathbb{R}^{d_\psi \times d_\psi}$, by stacking the columns of each $\hat{f}(\psi)$:

\begin{align}\label{eq:direct_sum_of_vectorization}
\textbf{f} = \bigoplus_{\psi \in \hat{G}}\text{vec}(\hat{f}(\psi)) = \begin{bmatrix}
\text{vec}(\hat{f}(\psi_1))   \\
  \text{vec}(\hat{f}(\psi_2))  \\
 \vdots \\
\end{bmatrix}  .
\end{align}

As stated before, Fourier coefficients has the equivariance property which can be expressed as $\widehat{g \cdot f}(\psi) = \psi(g)\widehat{f}(\psi)$. Given Eq. \eqref{eq:direct_sum_of_vectorization} and the equivariance property of the Fourier coefficients, vectorized function $[g \cdot f]$ can be rewritten as:

\begin{align}
\bigoplus_{\psi  \in \hat{G}}\text{vec}(\psi(g) \hat{f}(\psi)) &= \bigoplus_{\psi  \in \hat{G}} \Bigg( \bigoplus^{d_{\psi}} \psi(g) \Bigg) \text{vec}(\hat{f}(\psi)) \\
&= \bigoplus_{\psi \in \hat{G}} \Bigg( \bigoplus^{d_{\psi}}  \psi(g) \Bigg)\textbf{f} .
\end{align}

In simpler terms, the group $G$ acts on the vector $\textbf{f}$ with the following representation:
\begin{align}
    \rho(g) = \bigoplus_{\psi  \in \hat{G}}\bigoplus^{d_{\psi}} \psi(g).
\end{align} 

This means that $\rho(g) \mathbf{f}$ is the vector containing Fourier coefficients of the function $[g \cdot f]$. Representation $\rho$ acts on a space of functions over the group $G$. If $G$ is finite, this representation is \textbf{isomorphic} (equivalent) to the \textit{regular representation} defined in Sec. \ref{sec: Regular and Quotient nonlinearities}. Given $Q$ is the matrix that serves as a change of basis that performs Fourier transform, $Q^{-1}$ is the matrix that performs the inverse Fourier transform. This can be formulated as:
\begin{align}\label{eq:regular representation}
   \rho_{reg}(g) = Q^{-1} \Bigg( \bigoplus_{\psi  \in \hat{G}}\bigoplus^{d_{\psi}}  \psi(g) \Bigg) Q . 
\end{align}
where $d_{\psi}$ is the multiplicities of the corresponding irrep $\psi$, and also its dimension.

\subsection{Quotient Representation}\label{sec:quotient representation}
In Section \ref{sec:Inverse Fourier Transform}, we investigated the inverse Fourier transform on groups, and defined the final equation as:
\begin{align}\label{eq:appendix_inverse}
    f(g_i) = \hat{\delta}^T \rho(g_i)^T \hat{f}
\end{align}
where $\rho(g_i) = \bigoplus_{\psi \in \hat{G}} \bigoplus^{d_\psi} \psi(g)$ for regular representation and $\rho(g_i) = \bigoplus_{\psi \in \hat{G}} \psi(g)$ for quotient representation of $Q = SO(3)/SO(2)$. Recall that  $\hat{\delta} = \bigoplus_{\psi} \bigoplus^{q_\psi}\!\sqrt{d_\psi} vec(P_\psi)$ is interpreted as a vector with Fourier coefficients of a Dirac delta function centred at identity, and $\hat{f} = \bigoplus_{\psi}vec(\hat{f}(\psi))$ is a vector of Fourier coefficients of the irreps $\psi$. 

For the quotient representation $\rho$ with the quotient space $Q = SO(3)/SO(2)$, 
multiplication in $\hat{\delta}^T\rho(g_i)^T$ in Eq. \eqref{eq:appendix_inverse} results in only the mid-column of each irrep $\psi$ in $\rho(g_i)$. This can be illustrated as

\begin{align}
\rho(g_i)\hat{\delta} = 
\underbrace{
    \left(
        \begin{array}{c|c|c|c}
            \smallsquare \hspace{.5ex} \psi_0(g_i) &  & & \\
            \hline
             & \begin{array}{ccc}
\smallsquare\!\! & \!\!\!\smallsquare\!\!\! & \!\!\smallsquare \\[-1.85ex] 
\smallsquare\!\! & \!\!\!\smallsquare\!\!\! & \!\!\smallsquare \\[-1.85ex] 
\smallsquare\!\! & \!\!\!\smallsquare\!\!\! & \!\!\smallsquare
\end{array} \psi_1(g_i) & & \\
            \hline
             & & \begin{array}{ccccc}
\smallsquare\!\! & \!\!\smallsquare\!\! & \!\!\smallsquare\!\! & \!\!\smallsquare\!\! & \!\!\smallsquare \\[-2ex] 
\smallsquare\!\! & \!\!\smallsquare\!\! & \!\!\smallsquare\!\! & \!\!\smallsquare\!\! & \!\!\smallsquare \\[-2ex]
\smallsquare\!\! & \!\!\smallsquare\!\! & \!\!\smallsquare\!\! & \!\!\smallsquare\!\! & \!\!\smallsquare \\[-2ex]
\smallsquare\!\! & \!\!\smallsquare\!\! & \!\!\smallsquare\!\! & \!\!\smallsquare\!\! & \!\!\smallsquare \\[-2ex]
\smallsquare\!\! & \!\!\smallsquare\!\! & \!\!\smallsquare\!\! & \!\!\smallsquare\!\! & \!\!\smallsquare
\end{array} \psi_2(g_i) & \\
\hline
& & & \ddots
        \end{array}
    \right)
}_{\rho(g_i) = \bigoplus_{\psi \in \hat{G}} \psi(g_i)}
\cdot
\underbrace{
    \left(
    \scriptsize
        \begin{array}{c}
            \sqrt{d_{\psi_0}} \\
            \hline
            - \\
            \sqrt{d_{\psi_1}} \\
            - \\
            \hline
            - \\
            - \\
            \sqrt{d_{\psi_2}} \\
            - \\
            - \\
            \hline
            \vdots  
        \end{array}
    \right)
}_{\hat{\delta} }
\label{eq:rho_delta}
\end{align}

where $l$ in $\psi_l$ corresponds to the degree of spherical harmonics. Note that the blocks in $\hat{\delta}$ corresponds to the non-scaled matrices $\sqrt{d_\psi} vec(P_\psi)$ in Eq. \eqref{eq:final_inverse_fourier}.

Spherical harmonics are functions defined on the surface of a sphere that form an orthogonal basis for square-integrable functions on the sphere. When used in equivariant kernel design, they extend the feature space onto this spherical surface, allowing the kernel to capture more complex patterns.  Spherical harmonics can also be described as functions $\Tilde{Y}_m^l : SO(3) \rightarrow \mathbb{R}$ on $SO(3).$ and are considered as a special case of the Wigner D-matrices\footnote{Wigner D-matrices are unitary matrices representing irreducible representations of the groups $SU(2)$ and $SO(3)$}. Specifically, the $n=0$ order within these matrices corresponds to spherical harmonics, suggesting that spherical harmonics can be represented as the $n=0$ column of a Wigner-D matrix, formulated as
\begin{align}
    Y_m^l(\theta, \phi) = \frac{1}{\sqrt{2l+1}}D^l_{m0}(\theta, \phi, \gamma) .
\end{align}
This corresponds to $P_\psi$ selecting the columns of $\psi$ which are $H$ invariant in Eq. \eqref{eq:final_inverse_fourier}. The column, $D^l_{m0}$, consists of functions that are invariant to rotations around a particular axis, indicating that spherical harmonics remain invariant under such rotations. Consequently, focusing on the middle column of the Wigner-D matrices allows us to represent rotation-equivariant features in a more computationally efficient manner, as it requires fewer channels to be stored, while also maintaining efficiency \cite{bekkers2023fast}. Eq. \eqref{eq:rho_delta} illustrates how the middle column of the Wigner-D matrices are extracted in inverse Fourier transform.

\section{Nonlinear Layers}

\label{sec: other nonlinearities}

In equivariant networks, nonlinear layers must also satisfy the equivariance constraint. The nonlinearities discussed in this section operate on each feature vector $f(x) \in \mathbb{R}^{c_{\text{in}}}$ for all $x \in \mathbb{R}^n$. 

\paragraph{Norm Nonlinearity}
Point-wise nonlinearities acting on the norm of each field preserves the rotational equivariance, as described in \citet{harmonic_networks}. 
Norm nonlinearity can be formulated as:
\begin{align}\label{eq:norm_nonlinearity}
    f(x) \rightarrow \sigma\left(\|f(\mathbf{x})\|_2\right)\frac{f(\mathbf{x})}{\|f(\mathbf{x})\|_2}
\end{align}
where $\sigma$ represents the point-wise nonlinearity. In the case of $\sigma$ being ReLU, a learnable bias $ b \in \mathbb{R}^+$ can be introduced, which makes $\sigma\left(\|f(\mathbf{x})\|_2\right) = ReLU(\|f(\mathbf{x})\|_2 - b )$.

\paragraph{Gated Nonlinearity}
 \citet{3D_steerable_cnns} introduce gated nonlinearity, which can be considered as a special case of the norm nonlinearity. Gated nonlinearities act on a feature vector $f(\mathbf{x})$ by scaling its norm by a gating scalar that is computed via a sigmoid nonlinearity $\frac{1}{1+e^{-s(\mathbf{x})}}$, where s is a scalar feature field. In this case, $\sigma\left(\|f(\mathbf{x})\|_2\right)$ in Eq. \ref{eq:norm_nonlinearity} can be reformulated as $ \|f(\mathbf{x})\|_2 \frac{1}{1+e^{-s(\mathbf{x})}}$.

\paragraph{Tensor Product Nonlinearity}
Feature vectors $f_1(\mathbf{x}) \in \mathbb{R}^{c_1}$ and $f_2(\mathbf{x}) \in \mathbb{R}^{c_2}$ of arbitrary types $\rho_1$ and $\rho_2$ can be combined to a tensor product feature $(f_1 \otimes f_2)(\mathbf{x}) \in \mathbb{R}^{c_1c_2}$. Product output transforms under the representation $\rho_1 \otimes \rho_2 : G \rightarrow GL(\mathbb{R}^{c_1c_2})$. Tensor product operation is nonlinear and equivariant, therefore it can be integrated into the equivariant network without the need for any further nonlinearities. More details can be found in \citet{Clebsch_Gordan_Nets, 3D_steerable_cnns}.

\section{Equivariance of the Nonlinear Layer}\label{sec: Equivariance of the nonlinear layer}
In equivariant network architectures, the assumption is that the previous MLP and convolution layers are all equivariant, satisfying $\hat{f}(g.x) = \rho(g)\hat{f}(x)$. In our adaptive approach, the sampling matrix is generated through equivariant layers to ensure full equivariance, meaning that it satisfies the constraint
\begin{align}
    A(g.x) = A(x)\rho(g)^T    .
\end{align}

Given the equivariance of the previous layers and the sampling matrix $A$, equivariance of the activation layer can be shown as:
\begin{align}
\hat{f}(g.x) &= \frac{1}{N} A(g.x)^T \sigma(A(g.x) \hat{f}(g.x)) \\
\hat{f}(g.x) &= \frac{1}{N} (A(x) \rho(g)^T)^T \sigma(A(x) \rho(g)^T \rho(g) \hat{f}(x)) \\
\hat{f}(g.x) &= \rho(g) \frac{1}{N} A(x)^T \sigma(A(x) \hat{f}(x)) \\
\hat{f}(g.x) &= \rho(g) \hat{f}(x)
\end{align}

\section{Orthogonality Metrics}\label{sec:orthogonality metrics}
In Section \ref{sec:adaptive sampling matrix}, we discussed the orthogonality of the sampling matrix, meaning that it satisfies the condition $A^TA = AA^T = NI$, where $N$ is the number of group samples. In this section, we will compute the deviation of the sampling matrices from orthogonality using two metrics; $\epsilon_1$ in Eq. \eqref{eq:epsilon 1} and Eq. \eqref{eq:epsilon 2}. Note that the representations in this section are assumed to be \textbf{quotient representations}. Towards the end of this section, we briefly discuss how the metrics differ for regular representations. The first metric $\epsilon_1$ is defined as

\begin{align}\label{eq:epsilon 1}
    \epsilon_1 &= \frac{1}{N} \sum_{i,j < N} |A A^T - I_N|
\end{align}
where $N$ is the number of group samples, and $I_N$ is the identity matrix of size $N$. To simplify, $AA^T$ can be visualized as

\begin{align}
\hspace{-1.5cm}
\footnotesize 
\renewcommand{\arraystretch}{2} 
\setlength{\arraycolsep}{3pt} 
A A^T &= 
\underbrace{
\normalsize
    \left(
\begin{array}{c|c|c|c}
    \overset{\text{\raisebox{1mm}{$\psi_{0}(g_0)\hat{\delta}_0$}}}{\smallsquare} & \overset{\text{\raisebox{1mm}{$\psi_{1}(g_0)\hat{\delta}_1$}}}{\smallsquare\,\smallsquare\,\smallsquare}  & \cdots &  \overset{\text{\raisebox{1mm}{$\psi_{L}(g_0)\hat{\delta}_L$}}}{\smallsquare\, \smallsquare\, \cdots \smallsquare} \\
    \hline
    \vdots & \vdots & \ddots & \vdots \\
    \hline
    \overset{\text{\raisebox{1mm}{$\psi_{0}(g_N)\hat{\delta}_0$}}}{\smallsquare} & \overset{\text{\raisebox{1mm}{$\psi_{1}(g_N)\hat{\delta}_1$}}}{\smallsquare\,\smallsquare\,\smallsquare}  & \cdots  &  \overset{\text{\raisebox{1mm}{$\psi_{L}(g_N)\hat{\delta}_L$}}}{\smallsquare\, \smallsquare\, \cdots \smallsquare} 
\end{array}
\right) 
}_{A \hspace{.5em} \in \hspace{.5em} \mathbb{R}^{N \times F}} 
\cdot
\underbrace{
\footnotesize
    \left(
\begin{array}{c|c|c}
    \smallsquare \hspace{.25em} [\psi_{0}(g_0)\hat{\delta}_0]^T  
    & \cdots & \smallsquare \hspace{.25em} [\psi_0(g_N)\hat{\delta}_0]^T   \\
    \hline
    \begin{array}{c} \smallsquare \\ [-1.85ex] \smallsquare \\ [-1.85ex] \smallsquare \end{array} [\psi_{1}(g_0)\hat{\delta}_1]^T  
    & \cdots & \begin{array}{c} \smallsquare \\ [-1.85ex] \smallsquare \\ [-1.85ex] \smallsquare \end{array} [\psi_{1}(g_N)\hat{\delta}_1]^T    \\
    \hline
    \vdots & \ddots & \vdots \\
    \hline
    \begin{array}{c} \smallsquare \\  [-1.85ex] \smallsquare \\ [-2ex] \vdots \\[-1.75ex] \smallsquare \end{array} [\psi_{L}(g_0)\hat{\delta}_L]^T 
    & \cdots & \begin{array}{c} \smallsquare \\  [-1.85ex] \smallsquare \\ [-2ex] \vdots \\[-1.75ex] \smallsquare \end{array} [\psi_{L}(g_N)\hat{\delta}_L]^T 
\end{array}
\right) 
}_{A^T \hspace{.5em} \in \hspace{.5em} \mathbb{R}^{F \times N}}   \label{eq:AAT first }
\\ \nonumber
\\ 
&=
\underbrace{
\sum_{\psi \in \hat{G}} 
\large
    \left(
        \begin{array}{c|c|c}
            \overset{\text{\raisebox{1mm}{$ \psi(g_0)\hat{\delta} \cdot [\psi(g_0)\hat{\delta}]^T $}}}{\smallsquare}  & \cdots  & \overset{\text{\raisebox{1mm}{$  \psi(g_0)\hat{\delta} \cdot [\psi(g_N)\hat{\delta}]^T $}}}{\smallsquare}  \\
            \hline
            \vdots & \ddots & \vdots  \\
            \hline
            \overset{\text{\raisebox{1mm}{$ \psi(g_N)\hat{\delta} \cdot [\psi(g_0)\hat{\delta}]^T $}}}{\smallsquare}  & \cdots & \overset{\text{\raisebox{1mm}{$ \psi(g_N)\hat{\delta} \cdot [\psi(g_N)\hat{\delta}]^T $}}}{\smallsquare}
        \end{array}
    \right)
}_{AA^T \hspace{.5em} \in  \hspace{.5em} \mathbb{R}^{N \times N}} \label{eq:AAT second}
\end{align}

where $F$ is the size of the representation $\rho(g_i)\hat{\delta}$, which can be computed as $F = \sum_{l=0}^L 2l+1$ for $SO(3)$, with degree up to $L$. The matrices are formed in blocks to emphasize the multiplication between the corresponding representations. Each square $\begin{array}{c}\smallsquare \vspace{.25em} \end{array}$ in the matrix corresponds to a single matrix coefficient. 

Given the orthogonality of the compact group representations, $\rho(g^{-1}) = \rho(g)^{-1} = \rho(g)^T$, and the Peter-Weyl theorem (Thm. \ref{thm:orthonormal basis}),
 it is expected that the matrix product $AA^T$ yield an identity matrix. The $\epsilon_1$ quantifies the deviation of the representations from the perfect orthogonality.

Moreover, $A^TA$ can be illustrated as 

\begin{align}
\scriptsize
\renewcommand{\arraystretch}{1.4} 
\setlength{\arraycolsep}{3pt} 
A^TA &= \underbrace{
\footnotesize
    \left(
\begin{array}{c|c|c}
    \smallsquare \hspace{.25em} [\psi_{0}(g_0)\hat{\delta}_0]^T  
    & \cdots & \smallsquare \hspace{.25em} [\psi_0(g_N)\hat{\delta}_0]^T   \\
    \hline
    \begin{array}{c} \smallsquare \\ [-1.85ex] \smallsquare \\ [-1.85ex] \smallsquare \end{array} [\psi_{1}(g_0)\hat{\delta}_1]^T  
    & \cdots & \begin{array}{c} \smallsquare \\ [-1.85ex] \smallsquare \\ [-1.85ex] \smallsquare \end{array} [\psi_{1}(g_N)\hat{\delta}_1]^T    \\
    \hline
    \vdots & \ddots & \vdots \\
    \hline
    \begin{array}{c} \smallsquare \\  [-1.85ex] \smallsquare \\ [-2ex] \vdots \\[-1.75ex] \smallsquare \end{array} [\psi_{L}(g_0)\hat{\delta}_L]^T 
    & \cdots & \begin{array}{c} \smallsquare \\  [-1.85ex] \smallsquare \\ [-2ex] \vdots \\[-1.75ex] \smallsquare \end{array} [\psi_{L}(g_N)\hat{\delta}_L]^T 
\end{array}
\right) 
}_{A^T \hspace{.5em} \in \hspace{.5em} \mathbb{R}^{F \times N}} 
\cdot
\underbrace{
\normalsize
    \left(
\begin{array}{c|c|c|c}
    \overset{\text{\raisebox{1mm}{$\psi_{0}(g_0)\hat{\delta}_0$}}}{\smallsquare} & \overset{\text{\raisebox{1mm}{$\psi_{1}(g_0)\hat{\delta}_1$}}}{\smallsquare\,\smallsquare\,\smallsquare}  & \cdots &  \overset{\text{\raisebox{1mm}{$\psi_{L}(g_0)\hat{\delta}_L$}}}{\smallsquare\, \smallsquare\, \cdots \smallsquare} \\
    \hline
    \vdots & \vdots & \ddots & \vdots \\
    \hline
    \overset{\text{\raisebox{1mm}{$\psi_{0}(g_N)\hat{\delta}_0$}}}{\smallsquare} & \overset{\text{\raisebox{1mm}{$\psi_{1}(g_N)\hat{\delta}_1$}}}{\smallsquare\,\smallsquare\,\smallsquare}  & \cdots  &  \overset{\text{\raisebox{1mm}{$\psi_{L}(g_N)\hat{\delta}_L$}}}{\smallsquare\, \smallsquare\, \cdots \smallsquare} 
\end{array}
\right) 
}_{A \hspace{.5em} \in \hspace{.5em} \mathbb{R}^{N \times F}}  \label{eq:ATA first}\\
\nonumber
\\
&=
\normalsize
\underbrace{
   \sum_{g \in G}  \left(
   \scriptsize
    \begin{array}{c|c|c}
        \smallsquare \hspace{.3em}  \begin{array}{c}
        [\psi_0(g)\hat{\delta}_0]^T\psi_0(g)\hat{\delta}_0 \end{array}  &
        \cdots & \cdots \\[1ex]
        \hline
          \vdots  &  
          {\setlength{\arraycolsep}{.5pt} \scriptsize \begin{array}{ccccc} \smallsquare & & & & \\ [-2.5ex] & \smallsquare &  & & [\psi_1(g)\hat{\delta}_1]^T\psi_1(g)\hat{\delta}_1 \\[-1.5ex] & & \smallsquare &  & \end{array} } 
          & \cdots \\[1ex]
        \hline
        {\scriptsize \begin{array}{c} \smallsquare \\  [-1.85ex] \smallsquare \\ [-2ex] \vdots \\[-1.75ex] \smallsquare \end{array} 
[\psi_{0}(g)\hat{\delta}_0]^T \psi_{L}(g)\hat{\delta}_L }
        & \ddots & 
        {\setlength{\arraycolsep}{.5pt} \scriptsize 
        \begin{array}{ccccc} \smallsquare & &  & & 
        \\ [-2ex] & \ddots &  & &
        \\[-2ex] 
        &  & \smallsquare&  & \end{array} } \\
    \end{array}
    \right)
}_{A^T A \hspace{.5em} \in \hspace{.5em} \mathbb{R}^{F \times F}} \label{eq:ATA second}
\end{align}

Recall the orthogonality property of irreps in Peter-Weyl theorem (Thm. \ref{thm:orthonormal basis}). The orthonormality condition for the $\psi^l_{mn}(g)$ can be expressed as an integral over the group as in
\begin{align}
    \int_{SO(3)} \psi^{l}_{mn}(g) \psi^{l'}_{m'n'}(g)  \, d\mu = \delta_{ll'} \delta_{mm'} \delta_{nn'} \frac{1}{2l+1}
\end{align}
and in the discretized form
\begin{align}\label{eq:oo}
     \sum_{g \in G}  \psi^{l}_{mn}(g) \psi^{l'}_{m'n'}(g) = \frac{|G|}{2l+1} \delta_{ll'} \delta_{mm'} \delta_{nn'} 
\end{align}
where $\psi^l$ is an irrep of $SO(3)$ with degree $l$ and with $m, n$ matrix indices. $|G|$ is a finite number of group samples from $SO(3)$.

Given Eq. \eqref{eq:oo}, each square (\hspace{-.25em}$\begin{array}{c}\smallsquare \vspace{.25em} \end{array}$\hspace{-.25em}) in diagonal blocks $ \sum_{g \in G} [\psi_l(g)\hat{\delta}_l]^T\psi_l(g)\hat{\delta}_l$ (Eq. \eqref{eq:ATA second}) corresponds to $|G|$. Note that $2l+1$ in Eq. \eqref{eq:oo} is cancelled with $\hat{\delta}^T\hat{\delta}$ in the blocks (see $\hat{\delta}$ in Eq. \eqref{eq:rho_delta}). This makes the $A^TA$

\begin{align}
    A^TA = 
    \left(
    \begin{array}{c|c|c}
         |G|  &  &  \\
        \hline
            &  |G|   & \\
        \hline
        &  & \ddots \\
    \end{array}
    \right) \hspace{1.5em} \in \hspace{.25em} \mathbb{R}^{F \times F}.
\end{align}
To obtain the identity matrix, we can divide $A^TA$ by $|G|$. 

\begin{align}\label{eq:epsilon 2}
\epsilon_2 = \frac{1}{F} \sum_{i,j < d_{\rho}} \Big| \frac{1}{|G|} A^T A - I_{F} \Big|
\end{align}

The value $\epsilon_2$ helps us understand how closely the learned group representations align with the principles of the Peter-Weyl theorem.

\paragraph{Normalization}

The concepts discussed so far applies if $\hat{\delta}$ is not normalized in any way. Models in Sec. \ref{sec:Experiments} employ normalization over the vector $\hat{\delta}$ as in
\begin{align}\label{eq:delta normalization}
    \hat{\delta} \mapsto \Tilde{\delta} := \frac{\hat{\delta}}{\Vert \hat{\delta} \Vert }.
\end{align}  

Normalization over the entire vector has no effect on $\epsilon_1$, since in $AA^T$ (Eq. \eqref{eq:AAT second}), each matrix coefficient has the entire $\hat{\delta}$. However, in $A^TA$ (Eq. \eqref{eq:ATA second}), each matrix coefficient is associated with the respective segment of the kernel corresponding to the degree $l$. Therefore, in the computation of the $\epsilon_2$ (Eq. \eqref{eq:epsilon 2}), each matrix coefficient can be multiplied by the norm of the kernel to assess the orthogonality more accurately.
As an alternative, different normalization methods can be applied.

\paragraph{Regular Representation}

Regular representation is defined as
$$  \rho_{reg}(g) = Q^{-1} \Bigg( \bigoplus_{\psi}\bigoplus^{d_{\psi}}  \psi(g) \Bigg) Q .  $$
where $Q$ is the change of basis, $d_{\psi}$ is the size and the multiplicities of the corresponding irrep $\psi$. This means that, in the irreps decomposition, each irrep $\psi$ is repeated as many times as its size. For regular representations, $A \in \mathbb{R}^{N \times F}$ matrix takes a similar form, with the only difference being the repetition of the irreps, and the size $F = \sum_{l=0}^L (2l+1)^2 $.

\newpage

\section{Practical Design Choices}\label{sec:practical design choices}

\begin{wrapfigure}[45]{r}{0.38\textwidth}
\vspace{-1.5cm}
  \begin{center}
\begin{tikzpicture}[
  node distance=1cm and 0cm,
  mynode/.style={
    draw, rounded corners, text width=2cm, align=center,
    fill=#1, minimum height=1cm, minimum width=1.5cm
  },
  myarrow/.style={-Stealth, thick},
  mylabel/.style={text width=2cm, align=right},
  arrowlabel/.style={midway, fill=white, right, font=\small},
  arrowtext/.style={above, font=\small},
  rightnode/.style={draw,  rounded corners, text width=2cm, align=center, fill=#1, minimum height=1cm, minimum width=2cm}
]


\node[mynode=red!20] (block1) {\small{$k = 7$\\$p = 2$\\$s = 1$}};
\node[mylabel, left=of block1] (label1) {\small{Block 1}};
\node[mynode=red!20, below=of block1] (block2) {\small{$k = 5$\\$p = 2$\\$s = 1$}};
\node[mylabel, left=of block2] (label2) {\small{Block 2}};
\node[mynode=teal!20, below=of block2] (pooling1) {\small{$k = 5$\\$p = 1$\\$s = 2$}};
\node[mylabel, left=of pooling1] (label3) {\small{Pooling 1}};
\node[mynode=red!20, below=of pooling1] (block3) {\small{$k = 3$\\$p = 1$\\$s = 2$}};
\node[mylabel, left=of block3] (label4) {\small{Block 3}};
\node[mynode=teal!20, below=of block3] (pooling2) {\small{$k = 5$\\$p = 1$\\$s = 2$}};
\node[mylabel, left=of pooling2] (label5) {\small{Pooling 2}};
\node[mynode=red!20, below=of pooling2] (block4) {\small{$k = 3$\\$p = 2$\\$s = 2$}};
\node[mylabel, left=of block4] (label6) {\small{Block 4}};
\node[mynode=red!20, below=of block4] (block5) {\small{$k = 3$\\$p = 1$\\$s = 1$}};
\node[mylabel, left=of block5] (label7) {\small{Block 5}};
\node[mynode=teal!20, below=of block5] (pooling3) {\small{$k = 3$\\$p = 1$\\$s = 1$}};
\node[mylabel, left=of pooling3] (label8) {\small{Pooling 3}};
\node[mynode=red!20, below=of pooling3] (block6) {\small{$k = 1$\\$p = 0$\\$s = 1$}};
\node[mylabel, left=of block6] (label9) {\small{Block 6}};

\node[rightnode = red!20, right=1.5 cm of block2, yshift = 1.5cm] (newNode) {\small{$k = 1$\\$p = 0$\\$s = 1$}};
\node[mylabel, above=of newNode, right=1.2 cm of block2, yshift = 2.6cm ] (label10) {\small{\mbox{Sampling block}}};

\node[rightnode=blue!20, right=1.5cm of pooling1] (newNode2) {\small{$k = 5$\\$p = 1$\\$s = 4$}};
\node[mylabel, above=of newNode2, right=1.45cm of pooling1, yshift = 1cm] (label11) {\small{\mbox{Down \hspace{.15em} sampling}}};

\node[rightnode=blue!20, right=1.5cm of pooling2] (newNode3) {\small{$k = 3$\\$p = 0$\\$s = 2$}};
\node[mylabel, above=of newNode2, right=1.45cm of pooling2, yshift = 1cm] (label12) {\small{\mbox{Down \hspace{.15em} sampling}}};

\coordinate (aboveBlock1) at ([yshift=.2cm]block1.north);
\draw[myarrow] (aboveBlock1) -- (block1);
\node[arrowtext] at ([yshift=-0.5cm]aboveBlock1) {}; 

\coordinate (aboveBlock1) at ([yshift=.5cm]block1.north);
\coordinate (belowBlock6) at ([yshift=-.5cm]block6.south);

\draw[myarrow] (aboveBlock1) -- (block1);
\node[arrowtext] at ([yshift=0.1cm]aboveBlock1) {}; 
\draw[myarrow] (block1) -- (block2) node[arrowlabel] {};
\draw[myarrow] (block2) -- (pooling1) node[arrowlabel] {};
\draw[myarrow] (pooling1) -- (block3) node[arrowlabel] {};
\draw[myarrow] (block3) -- (pooling2) node[arrowlabel] {};
\draw[myarrow] (pooling2) -- (block4) node[arrowlabel] {};
\draw[myarrow] (block4) -- (block5) node[arrowlabel] {};
\draw[myarrow] (block5) -- (pooling3) node[arrowlabel] {};

\draw[myarrow] (pooling3) -- (block6) node[arrowlabel] {3};
\draw[myarrow] (block6) -- (belowBlock6);

\node[arrowtext] at ([yshift=-0.1cm]belowBlock6) {}; 
\draw[myarrow] ([yshift=1.2mm]block1.east) to[bend left] (newNode) ;   

\draw[myarrow] (newNode) to[bend left] ([yshift = -1mm]block1.east);


\draw[myarrow] (newNode) -- (newNode2) node[arrowlabel] {$A$};
\draw[myarrow] (newNode2) -- (newNode3) node[arrowlabel] {$A$};

\draw[myarrow] ($(newNode)!0.3!(newNode2)$) to[bend left] node[above, midway] {$A$} ([yshift=0mm]block2.east);

\draw[myarrow] ($(newNode2)!0.5!(newNode3)$) to[bend left] node[above, midway] {$A$} ([yshift=0mm]block3.east);

\coordinate (belowNewNode3) at ([yshift=-9cm]newNode3.south);
\draw[myarrow] (newNode3) -- (belowNewNode3) node[arrowlabel] {$A$};

\draw[myarrow] ([yshift=7.5cm]belowNewNode3) to[bend left] node[above, midway] {$A$} (block4.east);
\draw[myarrow] ([yshift=5cm]belowNewNode3) to[bend left] node[above, midway] {$A$} (block5.east);
\draw[myarrow] (belowNewNode3) to[bend left] node[above, midway] {$A$}(block6.east);
\end{tikzpicture}
  \end{center}
  \vspace{-.5cm}
  \caption{\footnotesize Architecture for voxel processing. The branch on the left represents the main branch that process the voxel data, while the branch on the right can be referred to as the sampling branch. The sampling block receives the first intermediate feature from the main branch and returns the sampling matrix $A$ to the first nonlinear layer in the first convolutional block. Each convolutional block comprises convolutional layer, batch normalization and activation layer.  } 
  \label{fig:architecture_mnist} 
\end{wrapfigure}
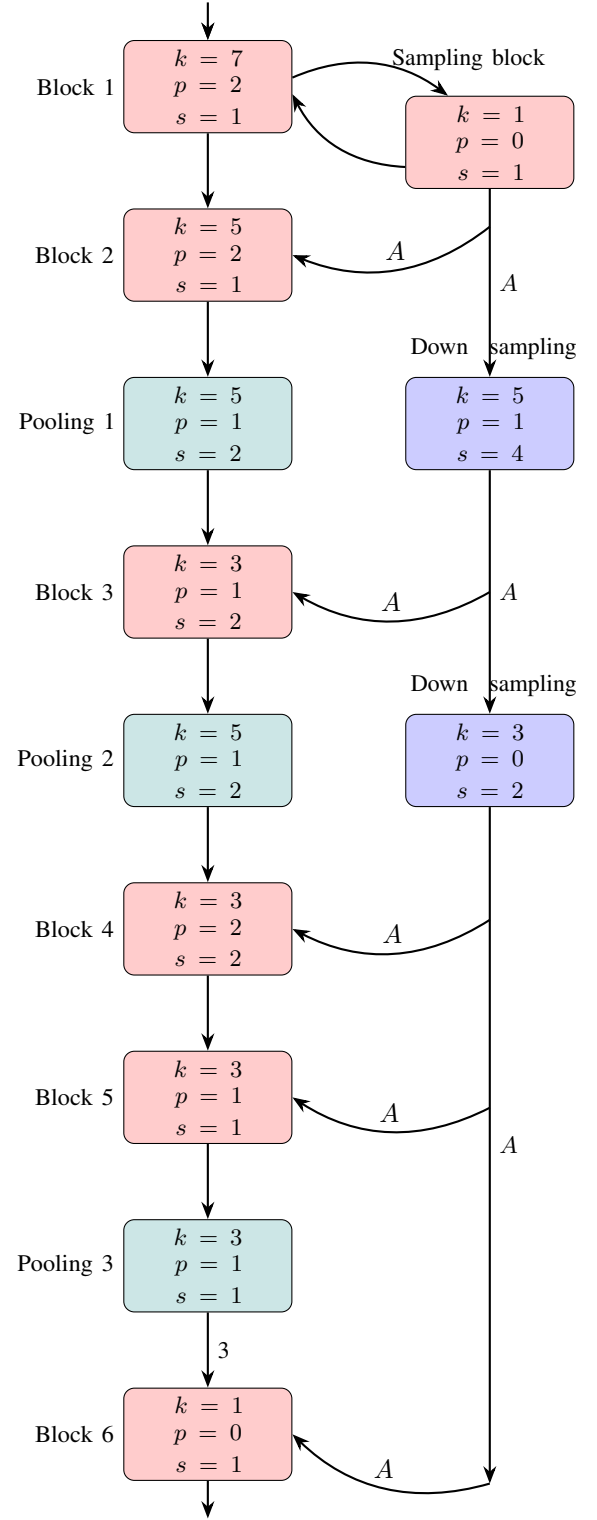

In Section \ref{sec:implementation}, we cover the implementation details, and here, we discuss various possible designs.

The implementation for voxel data is relatively straightforward by introducing convolutional layers for the generation and downsampling of the sampling matrix. For the point cloud processing, we have tried various implementation for the sampling branch by prioritizing the expressiveness and the computational efficiency.
In both architectures, the sampling matrix $A$ can be generated by processing the model input directly or by processing the first intermediate feature tensor, which corresponds to the output of the first pair of convolution and normalization layer. It is important to note that directly using the original input in an equivariant MLP may introduce specific challenges. The linear map $W$, in an equivariant MLP, is parameterized based on the Schur’s lemma, which states that the irrep intertwiners\footnote{An intertwiner is a linear map between two group \\
representations that commutes with the group action, \\ preserving the structure of the representations.} are zero for non-isomorphic\footnote{In group theory, an isomorphism is a bijective mapping \\ between two groups that preserves the group structure. \\ It respects the group operations and establishes a one-to-one \\correspondence between the elements of the groups.} irreps. In other words, if $\rho_1$ and $\rho_2$ are two non-isomorphic irreps, then any linear map $W$ such that $W\rho_1(g) = \rho_2(g)W \quad \forall g \in G$ must be the zero map. This property ensures that the space of intertwiners between different irreps is trivial unless the irreps are isomorphic. Model input is initially interpreted as feature space with representations of frequency zero. This means that $W$ will be parametrized in a way that the input won’t be mapped to a feature space with representations of higher frequencies, which results in information loss and less expressive intermediate features. This issue can be mitigated by the sampling branch processing the first intermediate features instead of model input, since the first feature space is generated by a convolutional layer which doesn’t have similar restrictions in the parametrization.

 Both point clouds and voxel data are downsampled through the convolutional layers in the main branch of the architecture. As detailed in (Sec. \ref{sec:implementation}), the model generates a unique sampling matrix for each point in the point clouds and each pixel in the voxel data. To ensure compatibility between the feature map and the sampling matrices, we apply subsampling to the sampling matrix. We can only employ convolutional layers in the processing of voxel data, however for point clouds, the simple approach would be to use indexing, where only the points that are still in the feature space are kept for further processing. This might result with the information loss since while the points are processed by message passing in the downsampling in the main architecture, the sampling matrix won't go through any further processing. Therefore, the expressiveness of the sampling matrices might decrease, particularly in the case where it is generated by an MLP layer which discards the neighbouring information.

 We have implemented multiple architectures given the points discussed. In Sec. \ref{sec:Experiments}, we only provide the results from the models that perform the best.

\newpage

\section{Experimental settings}
Here, we will outline the experimental settings and analyze the model results on two separate datasets. Additionally, we will present the findings in terms of model equivariance, runtime, and the orthogonality of the sampling matrices.

\paragraph{Equivariance}
 Due to the final invariant layer in the network, the entire network is expected to be invariant to rotations. To investigate the impact of different nonlinear layers and the number of group samples on the degree of equivariance in each model, we examined the relative equivariance error, which is defined as:
\begin{align}\label{eq:equivariance error}
    \epsilon_g(f) = \frac{\| f(x) - f(T(x)) \|}{\max \{ \|f(x)\|, \|f(T(x))\| \}}
\end{align}
where $f$ is the network, $T$ is the transformation of interest, which is $SO(3)$ rotations in this work, and $x$ is the model input.
It is important to note that this formulation computes the invariance error, since the numerator in the formulation corresponds to invariance. 
However, we will refer to it as the \textit{equivariance error}, to emphasize the degree of equivariance of the model before the final invariant layer empirically.

\paragraph{Orthogonality}

Recall from Sec. \ref{sec:adaptive sampling matrix} that our approach assumes the sampling matrix is orthogonal, satisfying \(AA^T = A^TA = NI\), where \(N\) is the number of group samples. In this section, we measure the deviation from orthogonality using two metrics introduced in Sec. \ref{sec:orthogonality metrics}. To compute the final \(\epsilon_1\) and \(\epsilon_2\) values (Eq. \eqref{eq:epsilons}), we calculate these metrics for each data sample and average the results over the samples in the test set.

\begin{align}\label{eq:epsilons}
\begin{aligned}
    \epsilon_1 &= \frac{1}{N} \sum_{i,j < N} (A A^T - I_N) \\
    \epsilon_2 &= \frac{1}{F} \sum_{i,j < d_{\rho}} \left( \frac{1}{|G|} (A^T A) - I_F \right)
\end{aligned}
\end{align}

In the context of matrix orthogonality, for a \( p \times q \) matrix, if \( p < q \), all rows can potentially be orthogonal and linearly independent. Conversely, if \( p > q \), not all rows can be orthogonal or independent due to dimensional constraints. In our specific case, the sampling matrix \( A \) has the shape \( N \times F \), where \( N \) represents the number of group samples and \( F \) is the size of the representations. This means that not all rows of \( A \) can be orthogonal to each other when the number of group samples exceeds the size of the representations.

\begin{wrapfigure}[12]{r}{0.55\textwidth}
    \centering   
    \vspace{-1cm}
\includegraphics[width=.55\textwidth]{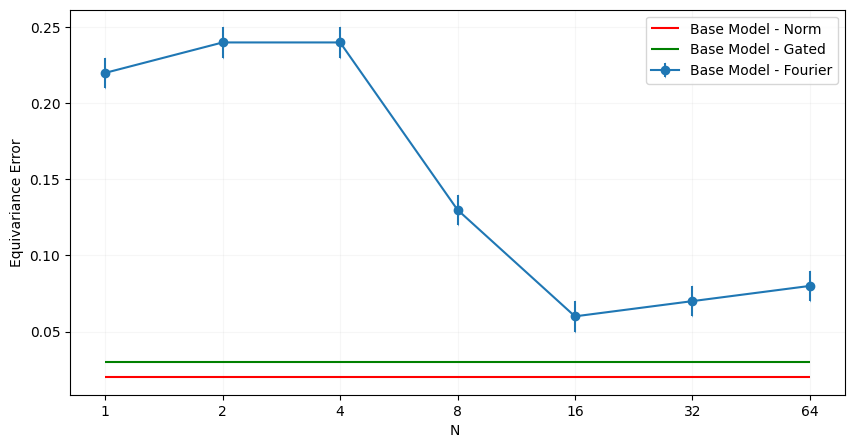}
\vspace{-.75cm}
    \caption{\small Equivariance error based on the number of group samples.}
    \label{fig:modelnet_equivariance}
\end{wrapfigure}

\subsection{ModelNet10}

To evaluate the model's robustness to SO(3) rotations, we generated a new set of training and test set. The new training set is generated by expanding the original ModelNet10 training set to three times its original size by incorporating random $SO(3)$ rotations into the point clouds. This means that each model is trained with potentially different datasets containing the same point clouds but in different orientations. The test set is also expanded to three times its original size by introducing $SO(3)$ rotations to the point clouds in the original dataset. However, the rotations for the test set are precomputed, ensuring consistency as all the models are evaluated on the same set. Training is conducted three times for each model, using three different seeds consistently to initialize the models. Each model is trained for 70 epochs and the highest test accuracy is reported in Fig \ref{fig:modelnet_accuracies}. All models discussed in this section uses Fourier-based nonlinearity with quotient representation. 

\paragraph{Equivariance}  Equivariance error is computed by rotating a single data sample by every octahedron rotations and taking the average of the errors with respect to each rotation. As seen in the Fig. \ref{fig:modelnet_equivariance}, the equivariance improves as the number of samples increases. In our approach, as described in Sec. \ref{sec: Equivariance of the nonlinear layer}, the network is designed to be fully equivariant. Consequently, the equivariance error is zero, which is consistent with our results.

\paragraph{Runtimes}

\begin{wrapfigure}[12]{r}{0.5\textwidth} 
  \centering
  \vspace{-.5cm}\includegraphics[width=0.5\textwidth]{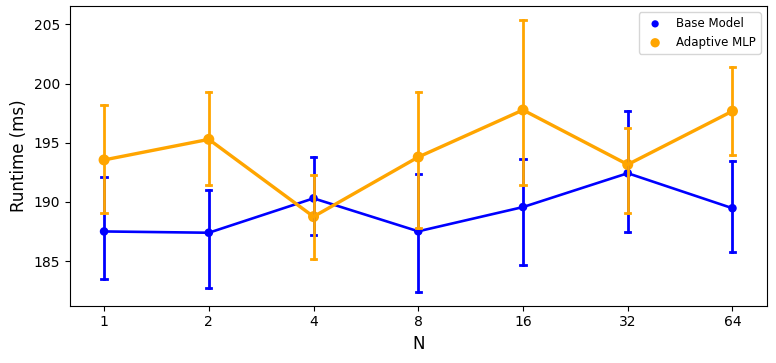} 
  \vspace{-.75cm}
  \caption{\small Inference runtime with respect to the number of group samples for Base model (blue) and Adaptive MLP (yellow).}
  \label{fig:modelnet_runtimes}
\end{wrapfigure}

The inference runtimes were obtained by averaging the results of eleven epochs, excluding the highest and lowest values among those runs. We also excluded the first ten batches for the warm-up.

The Adaptive MLP generally exhibits longer inference runtimes compared to the Base model, as reported in Fig. \ref{fig:modelnet_runtimes}. The only exception is when the model is trained with four samples. The increase in runtime for the adaptive approach can be attributed to the higher number of parameters (see Table \ref{table:modelnet_parameters}) and the additional computational steps involved in generating the sampling matrices and performing downsampling on them. Alternatively, using just-in-time (JIT) compilation could improve the results, but we did not experiment with that.

\begin{figure}[!h]
    \centering
    \includegraphics[width = .8\columnwidth]{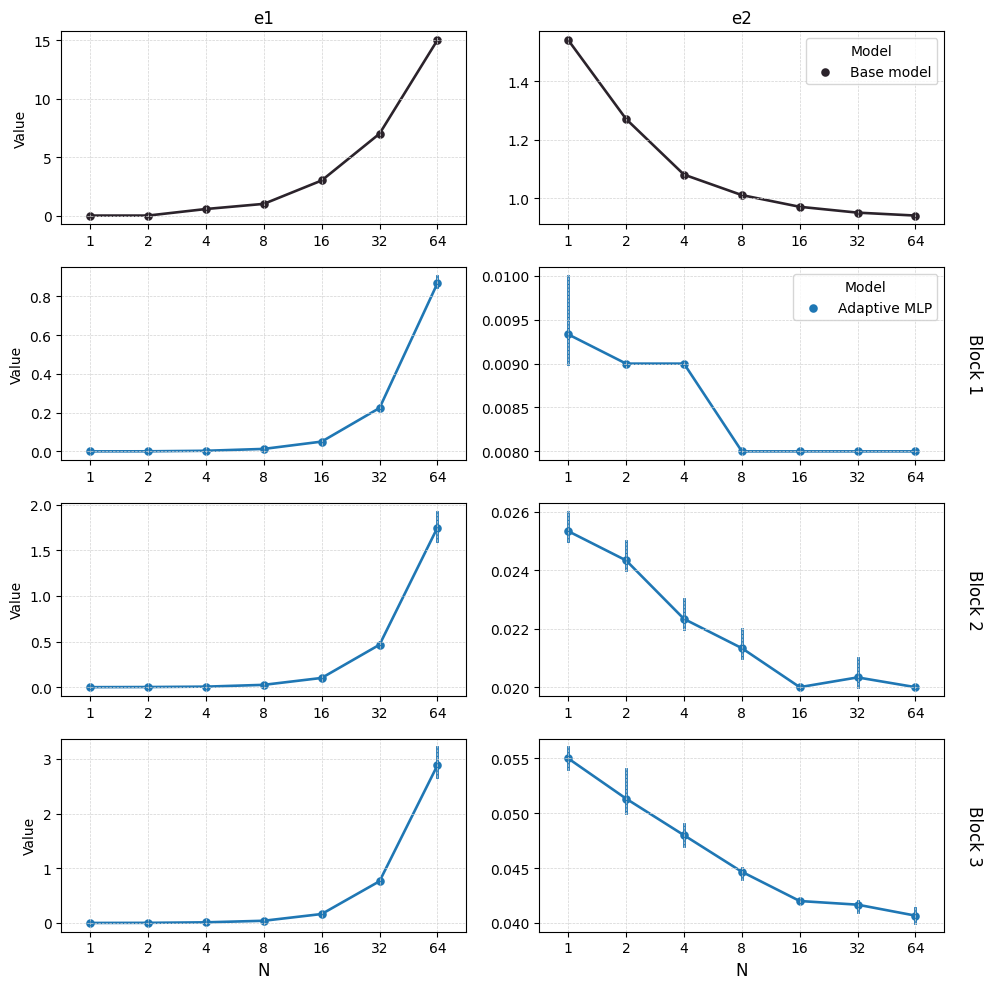}
    \caption{\small Orthogonality metrics for Base model and Adaptive MLP. The first row presents the values obtained from the Base model with respect to the number of group samples. Block 1, Block 2 and Block 3 refer to the sampling block, first downsampling block and the second sampling block, respectively, in the architecture (Fig. \ref{fig:architecture_modelnet}). Metrics are computed for the sampling matrices computed in each of those blocks, and presented on the second-to-last rows. }
    \label{fig:modelnet_orthogonality}
\end{figure}

\paragraph{Orthogonality} In the conventional approach, there is a single sampling matrix generated and used throughout the network. In our approach, Adaptive MLP generates a separate sampling matrix for each point in the point cloud, which enables to capture local symmetries. In this work, we use quotient representations for $Q=S^2\cong SO(3)/SO(2)$ up to degree $3$, where $F$ becomes equal to $\sum_{l=0}^{3} 2l+1 = 16$. This means that we can ideally achieve perfect orthogonality in terms of row independence, making $\epsilon_1$ close to zero as long as $N<16$. For $\epsilon_2$, the criteria for orthogonality is almost the opposite, meaning that we can obtain a value close to zero for  $\epsilon_2$ as long as $N>16$. This is indeed the case within the base models, as also shown in the first row of Fig. \ref{fig:modelnet_orthogonality}. Regarding the $\epsilon_1$, the values start to diverge around 8 samples, which is slightly earlier than expected. A similar pattern is observed for $\epsilon_2$. Notably, $\epsilon_2$ values do not converge to zero, but to around 0.7. This is due to normalization over the $\hat{\delta}$ in Eq. \eqref{eq:delta normalization}.

In our Adaptive MLP approach, we observe that even though we don't enforce orthogonality to the matrices, the network itself learns to create orthogonal sampling matrices. This is evident in both $\epsilon_1$ and $\epsilon_2$, which exhibit trends similar to those in the Base model, as shown in Fig. \ref{fig:modelnet_orthogonality}. Here, Block 1, Block 2, and Block 3 correspond to the sampling branch, first downsampling, and second downsampling blocks, respectively, as depicted in Fig. \ref{fig:architecture_modelnet}. $\epsilon$'s are computed for the sampling matrices processed in each one of those blocks. Additionally, from the second row onwards, there is a noticeable increase in $\epsilon_1$ and in $\epsilon_2$ from Block 1 to Block 3, particularly when $N > 16$. This trend further suggests that the model is autonomously learning orthogonality, without the need for external guidance. In terms of normalization applied to the sampling matrix, we experimented different techniques. However, these techniques didn't improve the model performance in terms of classification accuracy. The most effective normalization was the one applied over the entire $\hat{\delta}$ in Eq. \eqref{eq:delta normalization}.

\newpage

\subsection{NoduleMNIST3D}

MedMNIST3D \cite{medmnist} offers a collection of six biomedical image datasets, each containing 3D images standardized to 28x28x28 pixels and annotated with relevant classification label. In this work, the experiments are conducted on the MedMNIST3D Nodule dataset, which is a subset of the MedMNIST collection. This dataset is mainly designed for binary classification, focusing on the presence of the lung nodules in 3D CT scans. To evaluate the models, we generated rotated training, validation, and test sets. Rotating 3D voxels with random $SO(3)$ rotations introduces several issues such as interpolation artifacts due to misalignment with the original grid, and aliasing effects caused by the voxel grid's discrete nature. Furthermore, boundary effects at the grid edges can create artificial artifacts. To minimize the impact of those issues, we only perform cubic rotations, which is a subgroup of $SO(3)$, and we upsample each sample from 28 pixels to 29 pixels in each dimension.  

The training dataset is generated by applying random cubic rotations to each sample from the original training set, to ensure variation in the training process. The validation set is expanded to three times the size of the original test set by applying precomputed rotations. For the test set, we rotate each sample in the original test set with every cubic rotation, resulting in a set of size twenty-four times the original. We also trained the each model for 100 epochs and choose the one with the best validation set accuracy for testing. All results reported here are based on this test set performance. Furthermore, our implementation uses Fourier based nonlinearity with ELU as the pointwise nonlinearity and with regular representations up to degree $3$. All the hyperparameters used in the experiments are presented in Sec. \ref{sec:hyperparameters}. 

Additionally, in our experiments, we follow the approach, where the first intermediate features are used by the sampling branch to generate the sampling matrix. This approach is particularly effective since intermediate features, as opposed to the original input, already contain extracted features, and provide more information for the sampling branch. The full architecture of the model is illustrated in Fig. \ref{fig:architecture_mnist}.

\paragraph{Equivariance}

\begin{wrapfigure}[11]{r}{0.55\textwidth}
    \centering
    \vspace{-1cm}\includegraphics[width=.55\textwidth]{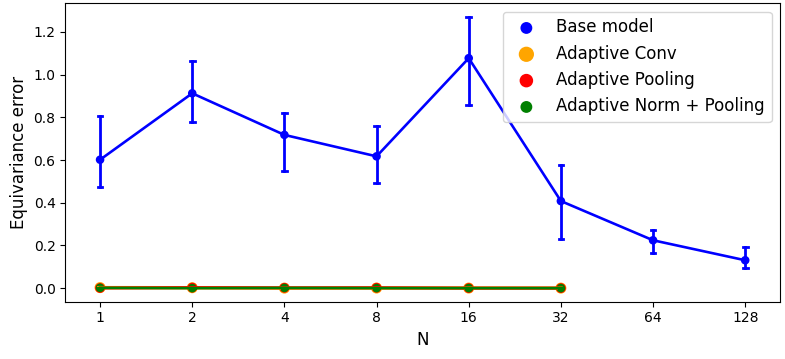}
    \vspace{-.75cm}
    \caption{\small Equivariance with respect to the number of samples } 
    \label{fig:mnist equivariance}
\end{wrapfigure}

Equivariance error is computed by applying all cubic rotations to a single Nodule data sample and then averaging the equivariance error over these operations. As shown in Fig. \ref{fig:mnist equivariance}, while the base model is not fully equivariant to rotations, the degree of equivariance tends to align with the number of group samples used, especially in models trained with sixteen or more samples. In contrast, our approach consistently ensures full equivariance, regardless of the implementation choice or the number of group samples.

\paragraph{Runtimes}

\begin{wrapfigure}[13]{r}{.55\textwidth}
    \centering
    \vspace{-.75cm}
    \includegraphics[width=.55\textwidth]{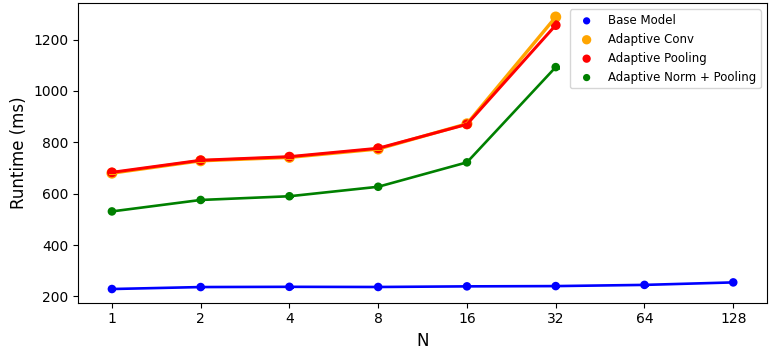}
    \vspace{-.75cm}
    \caption{\small Inference runtime with respect to the number of group samples by model. } 
    \label{fig:mnist_runtime}
\end{wrapfigure}

For consistency, each model was tested on the same test set over eleven epochs, and we calculated their average runtime, discarding the highest and lowest values from these runs. The results, depicted in Fig. \ref{fig:mnist_runtime}, represent the average runtime across models initialized with different seeds.

Fig. \ref{fig:mnist_runtime} reveals that our approach, incorporating extra layers into the architecture, typically results in an increased runtime compared to the base model. However, when the architecture's first nonlinearity is a norm nonlinearity, a shorter runtime is observed. This decrease in runtime can be attributed to the computationally less demanding nature of norm nonlinearities compared to the Fourier-based nonlinearities and the reduced number of parameters in the model, as detailed in Table \ref{table:mnist_parameters}.

\paragraph{Orthogonality}
Our approach generates a separate sampling matrix for each pixel in the voxel data. To minimize the computational complexity and also to obtain a more representative matrix, we first computed the average of the sampling matrices over the pixels for each sample and integrate the result into the nonlinear layer, to take a more global approach and also to reduce the computational cost to a certain degree. However, the final performance was quite low in terms of classification accuracy, therefore, we discard this implementation. In the computation of the final $\epsilon_1$ (Eq. \eqref{eq:epsilon 1}) and $\epsilon_2$ (Eq. \eqref{eq:epsilon 2}) for each model, we compute the metrics for each pixel in the data sample and average the results over the spatial dimensions, so that each data sample has a single $\epsilon_1$ and $\epsilon_2$ values. For the final results, we average each metric over the samples in the test set. 

\begin{figure}[!h]
    \centering
    \includegraphics[width = .8\columnwidth]{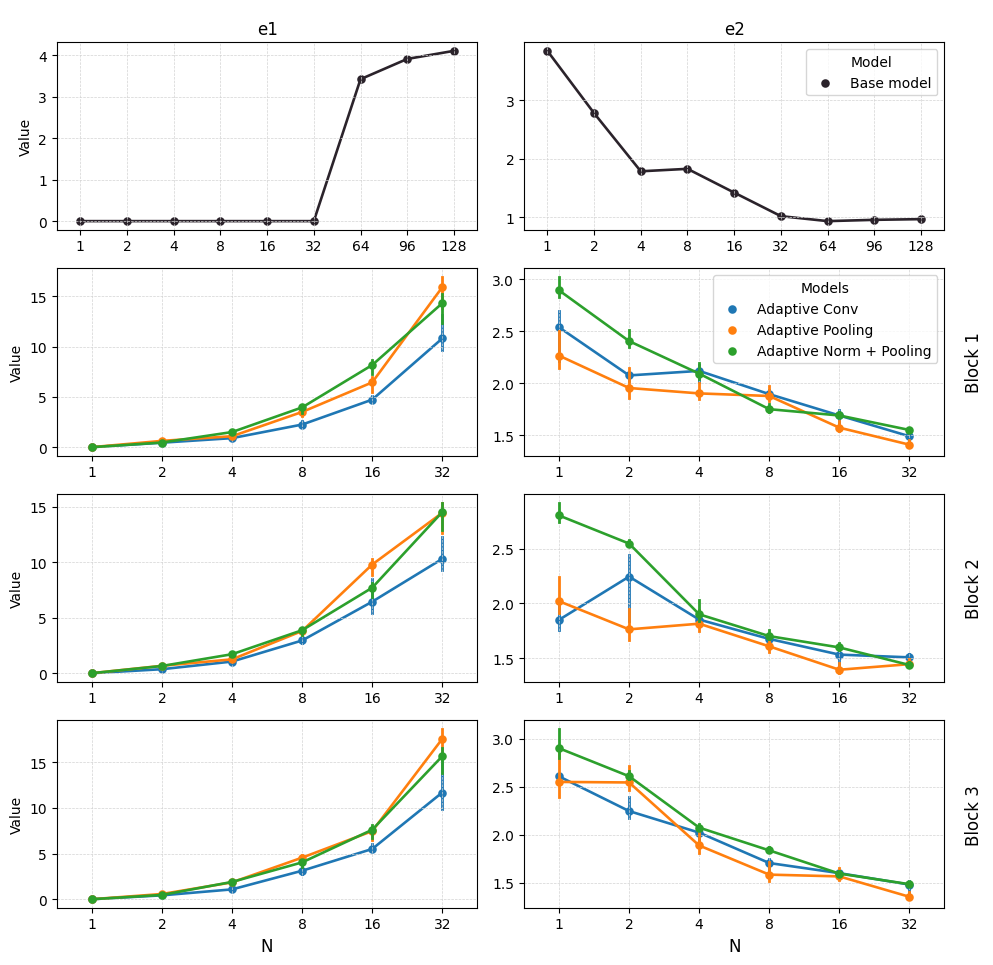}
    \caption{\small Orthogonality metrics for the Base model and Adaptive models. The first row presents the values obtained from the Base model. Block 1, Block 2, and Block 3 correspond to the sampling block, first downsampling, and the second downsampling block, respectively, in the architecture Fig. \ref{fig:architecture_mnist}.}
    \label{fig:mnist_orthogonality}
\end{figure}

In these experiments, we use regular representations up to and including degree 3, where $F$ becomes equal to $\sum_{l=0}^{3} (2l+1)^{2l+1} = 35$ (see Eq. \eqref{eq:regular representation}). This means that we can ideally achieve perfect orthogonality in terms of row independence, making $\epsilon_1$ close to zero as long as $N<35$, and making $\epsilon_2$ close to zero as long as $N>35$. This is indeed our observation, as also shown in the first row of Fig. \ref{fig:mnist_orthogonality}.  However, in the figure, see that the values are actually around 0.8, and not around zero. As previously noted in Sec. \ref{sec:orthogonality metrics}, the normalization of the sampling matrix impacts the computation of $\epsilon_2$. Given that the library by default normalizes across the entire vector of representations for each group sample, rather than on specific segments of the kernel representation, it is anticipated that $\epsilon_2$ may deviate slightly from zero. 

Similar to the results for ModelNet10, in our approach, the model demonstrates a tendency to approximate orthogonality in the sampling matrix, displaying a similar trend to the base model with respect to the number of group samples. The trend is more evident in $\epsilon_2$, where adaptive models tends to have a narrower error margin than the base model. We also experimented with various normalization methods for the sampling matrix. However, these modifications did not yield improvements in model performance regarding test accuracy. In the models we present, normalization is applied across the entire representation, similar to the approach used in the base model.

Additionally, it is important to note that while the base model was trained up to 128 group samples, we restricted our adaptive models to a maximum of thirty-two group samples. This was due to GPU memory constraints in our server, as models with higher number of group samples exceeded the available memory capacity.

\newpage

\section{Hyperparameters}\label{sec:hyperparameters}

\textbf{Point Cloud Processing}

\begin{minipage}{.55\textwidth}
    \centering
    \small
    \captionof{table}{Number of Parameters $(\times 10^4)$}
    \label{table:modelnet_parameters}
    \resizebox{\textwidth}{!}{
        \begin{tabular}{|c|c|c|c|c|c|c|c|}
            \hline
            
            Model & N1 & N2 & N4 & N8 & N16 & N32 & N64 \\ \hline
            Base model & \multicolumn{7}{c|}{93.97} \\ \hline
            Adaptive MLP & 94.35 & 94.4 & 94.49 & 94.69 & 95.07 & 95.82 & 97.35 \\ \hline
        \end{tabular}
    }
\end{minipage}
\hfill
\begin{minipage}{.45\textwidth}
    \centering
    \small
    \captionof{table}{Hyperparameters}
    \label{table:hyperparameters_modelnet}
    \renewcommand{\arraystretch}{1.25}
    \begin{tabular}{|l|l|}
        \hline
        \textbf{Hyperparameter} & \textbf{Value} \\ \hline
        batch size & 12 \\ \hline
        set abstraction ratio 1 & 0.190 \\ \hline
        set abstraction radius 1 & 0.542 \\ \hline
        set abstraction ratio 2 & 0.380 \\ \hline
        set abstraction radius 2 & 0.188 \\ \hline
        set abstraction ratio 3 & 0.475 \\ \hline
        set abstraction radius 3 & 0.368 \\ \hline
        width1 & 0.494 \\ \hline
        width2 & 0.339 \\ \hline
        width3 & 0.290 \\ \hline
        width4 & 2.000 \\ \hline
        number of rings & 5 \\ \hline
        learning rate & 0.0000087 \\ \hline
        weight decay & 0.088 \\ \hline
        frequency & 3 \\ \hline
        dropout & 0.100 \\ \hline
        channels conv & [16, 32, 64] \\ \hline
        channels mlp & [32, 64, 128] \\ \hline
        activation layer & Quotient \\ \hline
        maximum number of neighbors & 64 \\ \hline
    \end{tabular}
\end{minipage}

\vspace{0.5cm}

\textbf{Voxel Data Processing}

\begin{minipage}{.55\textwidth}
    \centering
    \small
    \captionof{table}{Number of Parameters $(\times 10^5)$}
    \label{table:mnist_parameters}
    \resizebox{\textwidth}{!}{
        \begin{tabular}{|c|c|c|c|c|c|c|c|}
            \hline
            Model & N1 & N2 & N4 & N8 & N16 & N32 & N64 \\ \hline
            Base model & \multicolumn{7}{c|}{15.398} \\ \hline
            Adaptive Conv & 15.4 & 15.43 & 15.5 & 15.8 & 17 & 21.7 & - \\ \hline
            Adaptive Pooling & 15.4 & 15.4 & 15.4 & 15.4 & 15.46 & 15.51 & - \\ \hline
            Adaptive Norm + Pooling & 14.72 & 14.73 & 14.85 & 14.78 & 14.83 & 14.9 & 15.17 \\ \hline
        \end{tabular}
    }
\end{minipage}
\hfill
\begin{minipage}{.45\textwidth}
    \centering
    \small
    \captionof{table}{Hyperparameters}
    \label{table:hyperparameters_voxel}
    \renewcommand{\arraystretch}{1.25}
    \begin{tabular}{|l|l|}
        \hline
        \textbf{Parameter} & \textbf{Value} \\ \hline
        batch size & 32 \\ \hline
        learning rate & 0.0001 \\ \hline
        activ & FourierELU \\ \hline
        frequency & 2 \\ \hline
        channels & [10, 20, 40, 60, 60, 60, 80] \\ \hline
        kernel sizes & [7, 5, 3, 3, 3, 1] \\ \hline
        paddings & [2, 2, 1, 2, 1, 0] \\ \hline
        strides & [1, 1, 2, 2, 1, 1] \\ \hline
    \end{tabular}
\end{minipage}

\end{document}